\documentclass[10pt,twocolumn,letterpaper]{article}

\usepackage[pagenumbers]{cvpr}

\usepackage[T1]{fontenc}

\usepackage{times}

\usepackage{graphicx}

\usepackage{amsmath,amssymb,bm}

\usepackage{booktabs,multirow,array,longtable,colortbl}

\usepackage{xcolor,enumitem,microtype}
\usepackage{placeins,stfloats}
\usepackage[colaction]{multicol}
\usepackage{fontspec}
\renewcommand{\tablename}{Table}

\definecolor{cvprblue}{rgb}{0.21,0.49,0.74}

\usepackage[pagebackref,breaklinks,colorlinks,citecolor=cvprblue,urlcolor=cvprblue,linkcolor=cvprblue]{hyperref}

\def\confName{CVPR}

\def\confYear{2027}

\graphicspath{{figures/}}

\newcounter{algorithm}

\setlist[itemize]{leftmargin=*,topsep=2pt,itemsep=1pt,parsep=0pt}

\title{MI-Hyper: Generating Multiple Instance Models for Weight Space Transfer Learning in Histopathology via Mixed Tasks Pretrained HyperNetwork}

\hypersetup{pdftitle={MI-Hyper: Generating Multiple Instance Models for Weight Space Transfer Learning in Histopathology via Mixed Tasks Pretrained HyperNetwork},pdfauthor={Xitong Ling, Yousef Kotp, Weiming Chen, Yining Li, Jiawen Li, Minxi Ouyang, Xinyi Guo, Mingxi Fu, Yizhi Wang, Tian Guan, Lianghui Zhu, Yonghong He}}

\makeatletter
\renewcommand{\@maketitle}{%
  \newpage
  \null
  \begin{center}
    {\Large\bfseries\@title\par}
    \bigskip
    {\large\begin{tabular}[t]{c}\@author\end{tabular}\par}
    \smallskip
  \end{center}
}
\makeatother
\author{
  {\fontsize{11.8}{14.2}\selectfont Xitong Ling\textsuperscript{1,}\thanks{These authors contributed equally.}, Yousef Kotp\textsuperscript{2,*}, Weiming Chen\textsuperscript{1}, Yining Li\textsuperscript{1}, Jiawen Li\textsuperscript{1}, Minxi Ouyang\textsuperscript{1}}\\
  {\fontsize{11.8}{14.2}\selectfont Xinyi Guo\textsuperscript{3}, Mingxi Fu\textsuperscript{1}, Yizhi Wang\textsuperscript{1}, Tian Guan\textsuperscript{1,}\thanks{Corresponding authors.}, Lianghui Zhu\textsuperscript{1,\textdagger}, Yonghong He\textsuperscript{1,\textdagger}}\\
  {\fontsize{10.2}{12.2}\selectfont \textsuperscript{1}Shenzhen International Graduate School, Tsinghua University}\\
  {\fontsize{10.2}{12.2}\selectfont \textsuperscript{2}Concordia University, Montréal, Canada\quad
   \textsuperscript{3}South China Normal University}
}

\begin{document}

\raggedbottom

\setlength{\floatsep}{13pt}

\setlength{\textfloatsep}{16pt}

\setlength{\intextsep}{13pt}

\maketitle
\suppressfloats[t]

\begin{abstract}
Zero-shot whole-slide image (WSI) prediction aims to generalize to new pathology tasks without parameter updates at deployment. MI-Zero makes these predictions by pooling patch--text similarities. Conditional weight generation, combined with supervision from diverse pathology tasks, offers an alternative: learning to construct predictive rules from task descriptions. We introduce MI-Hyper, a hypernetwork pretrained across mixed tasks to generate the aggregation and classification weights of multiple instance learning (MIL) models. We first train the generator and a shared MIL basis jointly using source labels. We then freeze the basis and refine the generator. The generated task weights are reused across cases. Across 308 tasks from 52 pathology datasets, MI-Hyper improves mean AUC, balanced accuracy and macro-F1 over MI-Zero by 1.85, 1.17 and 0.57 percentage points, respectively. On some molecular prediction tasks, both AUC and balanced accuracy improve by more than 20 points. These results support task-conditioned weight generation as a promising approach to weight-space transfer for long-sequence modeling.

\end{abstract}

\begin{figure}[t]

\centering

\includegraphics[width=\columnwidth]{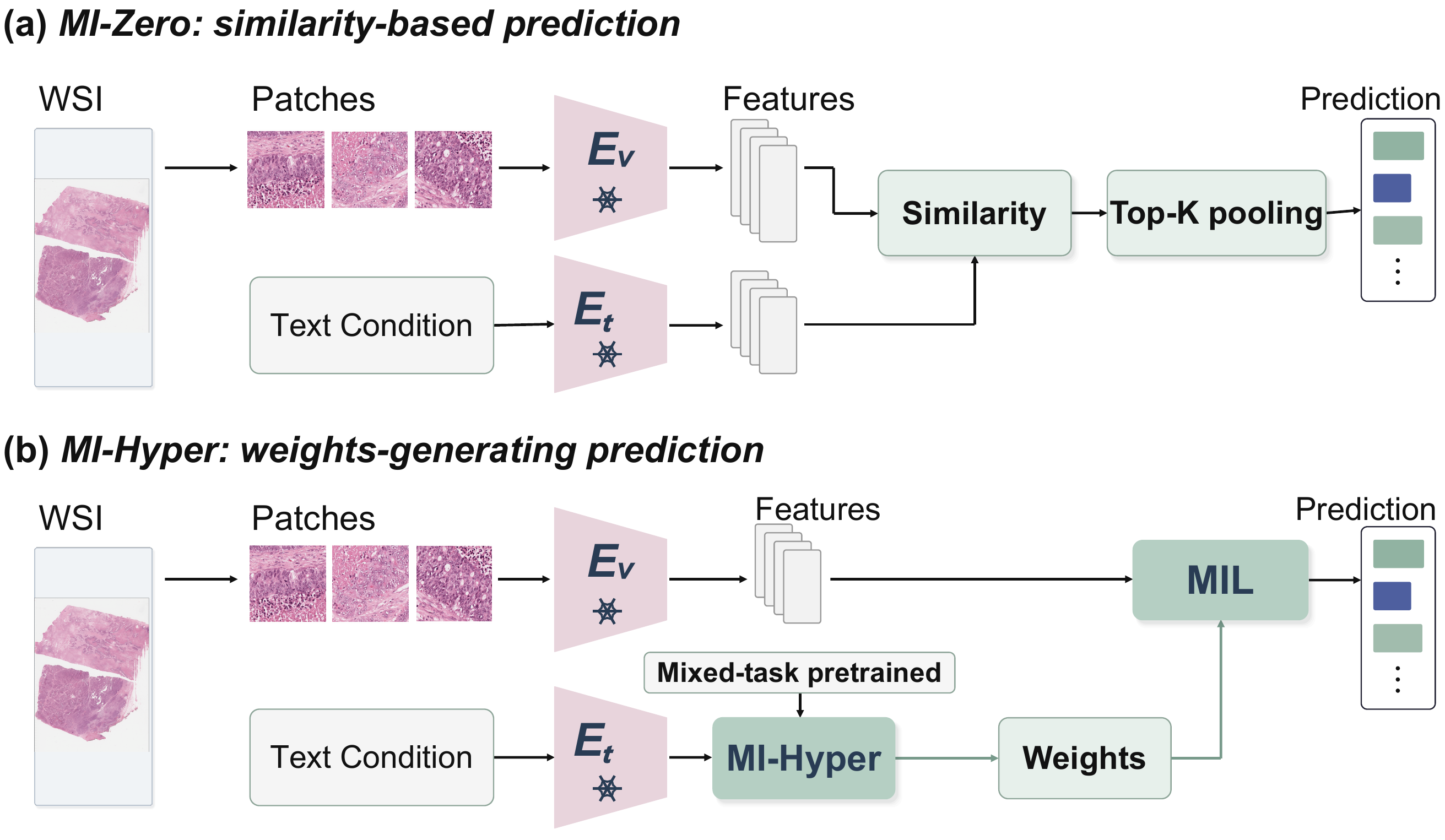}

\caption{\textbf{Constructing predictors from tiled WSIs.} (a) MI-Zero pools patch--text similarity scores. (b) MI-Hyper constructs an MIL predictor using generated weights.}

\label{fig:concept}

\end{figure}

\section{Introduction}
\label{sec:intro}

Zero-shot whole-slide image (WSI) prediction seeks to reuse a trained system for new pathology tasks without task-specific optimization at deployment. Because the same slide can support subtyping, grading and molecular prediction, the relevant tissue evidence depends on the endpoint. A zero-shot predictor must therefore select and combine patterns across long patch sequences according to the task.

Pathology foundation models provide a common starting point through reusable representations at the patch and whole-slide levels~\cite{lu2024conch,ding2025titan}. For patch-based prediction, multiple instance learning (MIL) converts a bag of representations into a slide-level output through aggregation and classification~\cite{ilse2018attention,lu2021clam}. Aggregation and classification parameters determine how tissue features contribute to a particular endpoint. The remaining question is how to obtain these parameters for a new task without additional target-task optimization.

Language provides a natural interface for specifying tasks through class descriptions. MI-Zero uses this interface to compare patch and class-text representations in a shared embedding space, then pools their similarities into slide-level predictions~\cite{lu2023mizero}. Changing the descriptions changes the similarity scores, which a predefined operator pools. This language interface also suggests a complementary route: generating the parameters that govern MIL aggregation and classification from the descriptions.

Conditional weight generation provides a mechanism for learning this mapping from descriptions to predictors. Hypernetworks generate model parameters from conditioning information, and language-conditioned methods extend this principle to classifiers and adaptation weights~\cite{ha2017hypernetworks,amosy2024text2model,chen2025texttolora}. Meanwhile, labeled pathology datasets cover multiple organs and prediction endpoints~\cite{yan2025pathorchestra,zhu2026digepath}. These source labels can supervise predictions made by generated models, allowing the description-to-weight mapping to be learned across tasks. This provides a basis for studying transfer through the weights of generated MIL predictors.

Building on this opportunity, we introduce MI-Hyper, a hypernetwork pretrained across mixed tasks to construct MIL predictors from task and class descriptions. Figure~\ref{fig:concept} contrasts MI-Zero's similarity pooling with MI-Hyper's weight generation, while Figure~\ref{fig:architecture} presents the training and deployment framework. We first train the generator and a shared MIL basis through source-label prediction. With the basis fixed, we then refine the dual-text generator using source labels and text-consistency objectives. At deployment, MI-Hyper generates task-specific aggregation and classification weights for each text condition and reuses them across cases. The generated parameters are shared within a task, but attention is computed from each case's tissue content. Their number is independent of bag length, so task-level model generation is separate from computation over each case's complete patch sequence. This separation allows us to study weight-space transfer in long-sequence modeling.

We train MI-Hyper with five folds over 112 semantic groups and compare it with MI-Zero on 308 tasks from 52 pathology datasets. Target task IDs are held out from source optimization and checkpoint selection, while cases can recur across tasks. On the complete query set, MI-Hyper improves mean AUC, balanced accuracy and macro-F1 by 1.85, 1.17 and 0.57 percentage points, respectively. On some molecular prediction tasks, both AUC and balanced accuracy improve by more than 20 points. To understand these results, we organize performance by endpoint, anatomy, bag size and dataset source, then examine individual tasks and the relation between generated queries and spatial attention.

Our contributions are threefold:
\begin{itemize}[label=$\bullet$]
\item We introduce MI-Hyper, which learns to generate task-specific MIL aggregation and classification weights from descriptions, separating predictor construction from case-level sequence processing.
\item We train the generator on labels from mixed source tasks through joint learning and fixed-basis refinement. The generated parameters are reused across cases, while chunked execution processes each complete patch sequence.
\item We evaluate this framework on 308 tasks from 52 pathology datasets, combining overall and task-level comparisons with analyses of generated queries and spatial attention.
\end{itemize}

\section{Related Work}
\label{sec:related}
\paragraph{Pathology foundation models and MIL.}
WSI analysis spans cancer detection and molecular prediction, with reliable evaluation depending on accurate reference labels~\cite{campanella2019clinical,coudray2018classification,kather2019msi,ling2025camelyonplus}. Large-scale self-supervision, exemplified by UNI and Virchow, and contrastive learning produce reusable patch representations~\cite{chen2024uni,vorontsov2024virchow,wang2022ctranspath,wang2023retccl,zimmermann2024virchow2}. Representation learning also encompasses cell-level morphology~\cite{chen2026beyondvit}, global slide context~\cite{chen2022hipt,wang2024chief,xu2024gigapath} and multimodal learning from reports or molecular measurements~\cite{ding2025titan,shaikovski2024prism,zhang2026care}. To obtain slide predictions from patch features, ABMIL uses attention, CLAM uses clustering-constrained attention, and TransMIL models correlations among instances~\cite{ilse2018attention,lu2021clam,shao2021transmil}. Related variants incorporate dual streams, agent tokens, graph interactions and distillation~\cite{li2021dsmil,ling2024amdmil,fu2026deformablegraph,li2024wikg,zhang2022dtfd}. Conventional task-supervised MIL fits aggregation and classification parameters to each downstream task. MI-Hyper instead generates these parameters from task descriptions.

\paragraph{Language-guided prediction.}
Visual--semantic and contrastive learning align images with textual descriptions~\cite{reed2016descriptions,radford2021clip}. Learnable and image-conditioned prompts further adapt language-guided prediction to general visual recognition~\cite{zhou2022coop,zhou2022cocoop}. In biomedicine, pathology-specific models such as CONCH and broader biomedical models learn image--text alignment from paired data~\cite{huang2023plip,lu2024conch,javed2024cplip,zhang2023biomedclip}. At the slide level, reports and clinical dialogue provide language supervision for attention and representation learning~\cite{shao2026limit,ding2025titan,vorontsov2026prism2}. For zero-shot WSI prediction, MI-Zero pools similarities between patches and text~\cite{lu2023mizero}. Our MI-Zero implementation uses the same frozen image and text encoders as MI-Hyper. It applies CONCH-Z's contrastive visual projection before pooling patch--text similarities, whereas MI-Hyper uses the patch features for MIL prediction.

\paragraph{Conditional weight generation.}
Early zero-shot recognition methods predicted classifiers and convolutional filters from text~\cite{elhoseiny2013write,ba2015predicting}. More generally, hypernetworks generate model parameters from conditioning information~\cite{ha2017hypernetworks,babu2024hyperfields}. This mechanism generates adapters and prompts~\cite{karimimahabadi2021hyperformer,he2022hyperprompt}, supports instruction conditioning~\cite{ivison2023hint}, and can produce low-rank adaptation weights based on LoRA~\cite{hu2022lora,chen2025texttolora}. Language can also guide adaptation in a hypernetwork's latent space~\cite{nava2023metaguidance} or determine an image encoder's normalization parameters~\cite{akinwande2026hyperclip}. Set representations encode unordered class descriptions~\cite{zaheer2017deepsets,lee2019settransformer}, while permutation-equivariant prediction preserves their correspondence with class outputs~\cite{amosy2024text2model}. MI-Hyper studies this description-to-weight mapping for MIL predictors operating on long WSI patch sequences.

\section{Methods of MI-Hyper}
\label{sec:method}

\begin{figure*}[t]
\centering
\includegraphics[width=\textwidth]{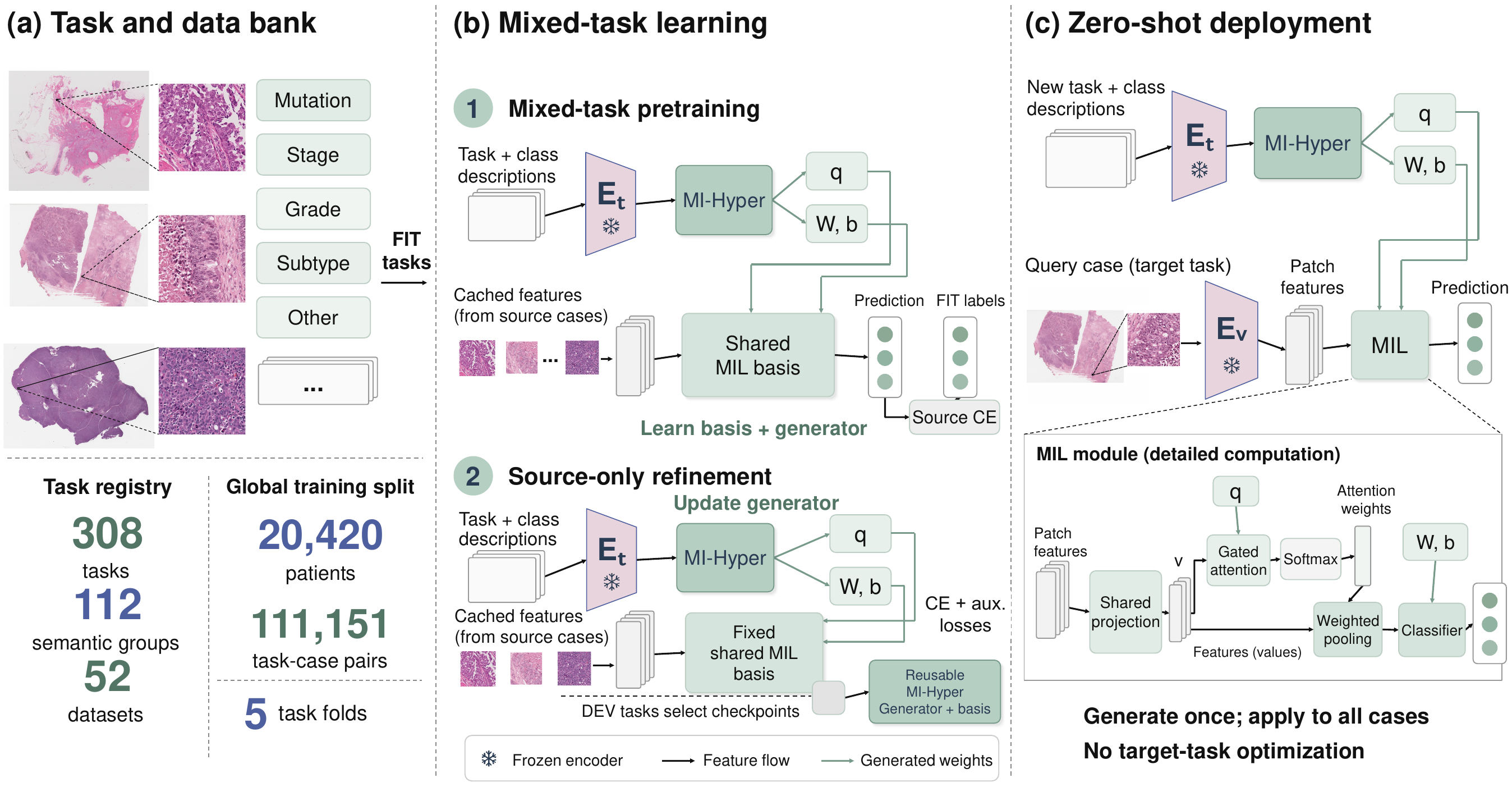}
\caption{\textbf{Mixed-task learning and zero-shot deployment of MI-Hyper.} (a) The task registry covers 52 datasets, 308 tasks and 112 semantic groups. The global training partition contains 20,420 patients and 111,151 task--case pairs. Each fold optimizes on eligible source tasks and holds out target task IDs; cases may recur across source and target tasks. (b) Joint source-label learning is followed by fixed-basis refinement with cross-entropy and auxiliary losses (Eq.~\eqref{eq:mihyper-refine}). (c) Task and class descriptions generate the aggregation query and classifier weights, which are reused across cases without target-task parameter updates.}
\label{fig:architecture}
\end{figure*}

Figure~\ref{fig:architecture} shows how source labels supervise generated MIL models through two-stage training. At deployment, each predictor reuses its generated parameters across cases and aggregates each case's complete patch sequence.

\subsection{Problem Formulation}

We represent a case as a bag of patch features extracted by a frozen encoder, $X=\{x_i\}_{i=1}^{N}$, where $x_i\in\mathbb R^d$. A bag may contain patches from one or more slides of the same case, and its length $N$ varies with tissue coverage. A prediction task $\tau$ is defined by an ordered vocabulary of $C_\tau$ classes and a text condition $c_\tau$ comprising class descriptions and additional task context.

Given $c_\tau$, the hypernetwork generates task-specific parameters and assembles them with shared components to form an MIL predictor:
\begin{align}
 \theta_\tau&=H_\phi(c_\tau),\qquad
 \mathcal W_\tau=\mathcal A_\nu(W_0,\theta_\tau),\nonumber\\
 p_\tau(y\mid X)&=f_\nu(X;\mathcal W_\tau).
 \label{eq:generate}
\end{align}
Here, $\phi$ denotes the generator parameters, $W_0$ contains components shared across tasks, and $\nu$ specifies the predictor parameterization and assembly rule. We instantiate this formulation with a gated MIL predictor.

Prior hypernetwork-based transfer for WSI prognosis uses WSI-conditioned mixing and sparse aggregation of cancer-specific task vectors~\cite{liu2026steph}. MI-Hyper instead conditions its generator on task and class descriptions, allowing the generated parameters to be reused across cases of the same task.

\subsection{Conditional Weight Generation}

Our gated MIL implementation uses frozen CONCH v1.5 patch features and CONCH-Z text embeddings, both of width 768~\cite{shao2026limit}. Class names specify the required distinctions, and background descriptions place the prediction endpoint in context. A learnable adapter then maps normalized text embeddings to 256-dimensional representations. For task context $s_\tau$ and class description $s_{\tau j}$,
\begin{align}
 t_\tau&=g_\phi(e(s_\tau)),\qquad
 z_{\tau j}=g_\phi(e(s_{\tau j})),\nonumber\\
 \bar z_\tau&=\frac{1}{C_\tau}\sum_{j=1}^{C_\tau}z_{\tau j},
 \label{eq:mihyper-text}
\end{align}
where $e$ is the frozen text encoder and $g_\phi$ is the trainable adapter. 
The class-name branch generates the attention query from task context and a symmetric class summary. The shared class decoder also receives the individual class representation and outputs its classifier row and bias:
\begin{align}
 q_\tau&=D_\phi^{q}([t_\tau;\bar z_\tau]),\nonumber\\
 (w_{\tau j},b_{\tau j})&=
 D_\phi^{\mathrm{cls}}([t_\tau;z_{\tau j};\bar z_\tau]).
 \label{eq:mihyper-decode}
\end{align}
The query has width $a=128$, and each classifier row has width $h=256$. Stacking the class outputs gives $W_\tau^{\mathrm{cls}}\in\mathbb R^{C_\tau\times h}$ and $b_\tau\in\mathbb R^{C_\tau}$. The symmetric class summary makes the query invariant to class order, while classifier rows follow the order of their descriptions. The shared decoder therefore accommodates tasks with different class counts.

The dual-text generator extends the class-name branch with a joint class-name--morphology branch. This branch has a separate trainable text adapter for morphology descriptions and a fusion adapter that combines each class's name and morphology representations. An availability mask excludes missing or unverified morphology descriptions before class summarization and nonlinear interaction. A text-conditioned gate then linearly interpolates the queries, classifier rows and biases of the two branches, component by component. The gate receives task and class text representations, pairwise similarity statistics within and between the name and morphology embeddings, and morphology availability. It does not receive patch features from the query case. Text therefore determines the task weights before case features enter MIL prediction.

\subsection{MIL Prediction}

To apply the generated parameters, the shared MIL basis first projects patch features into a common representation space and computes gated attention features:
\begin{align}
 v_i&=\operatorname{ReLU}(Px_i+b_P),\nonumber\\
 g_i&=\tanh(Vv_i+b_V)\odot
 \operatorname{sigmoid}(Uv_i+b_U).
 \label{eq:mihyper-basis}
\end{align}
The projection and gated attention parameters belong to the shared basis $W_0$. The generated query then scores each patch to obtain globally normalized attention and an aggregated case representation:
\begin{align}
 r_i&=q_\tau^\top g_i,\nonumber\\
 \alpha_i&=\frac{\exp(r_i)}{\sum_{j=1}^{N}\exp(r_j)},\qquad
 v_X=\sum_{i=1}^{N}\alpha_i v_i.
 \label{eq:pooling}
\end{align}
The generated classifier maps this representation to class probabilities:
\begin{equation}
 p_\tau(y\mid X)=\operatorname{softmax}
 \left(W_\tau^{\mathrm{cls}}v_X+b_\tau\right).
 \label{eq:mihyper-classifier}
\end{equation}
All classes share one task-conditioned attention distribution, so each bag is scored and pooled once per text condition. Although $q_\tau$, $W_\tau^{\mathrm{cls}}$ and $b_\tau$ are reused across cases, attention scores are recomputed from the patches of each case. We center the classifier rows and biases along the class dimension to remove a common logit offset. Since all classes use the same aggregated representation, this operation preserves softmax probabilities.

\subsection{Mixed-Task Learning}

Source optimization tasks (SOURCE-FIT) supply labeled cases, while separate source tasks (SOURCE-DEV) select checkpoints. Neither role includes the target task IDs for that fold. For a source task $\tau$, text condition $c_\tau$ and case--label pair $(X,y)$, the prediction objective is
\begin{equation}
 \mathcal L_{\mathrm{src}}(W_0,\phi)=
 \mathbb E_{(\tau,c_\tau,X,y)\sim\mathcal D_{\mathrm{src}}}
 \left[-\log p_\tau(y\mid X)\right].
 \label{eq:source-loss}
\end{equation}
Training cycles through source tasks, balances classes within each task and samples at most 2048 stored patches per case. Both pretrained encoders remain frozen.

Source cross-entropy jointly trains a randomly initialized generator and shared MIL basis; the pretrained encoders remain frozen. This stage uses canonical task contexts and class-specific morphology descriptions. SOURCE-DEV selects checkpoints by task-average AUC under the canonical condition, without parameter updates. The selected generator initializes the class-name branch, and a copy of its text adapter initializes the morphology branch. Refinement fixes the projection and gated attention basis $W_0$.

During fixed-basis refinement, each source case is evaluated using a canonical text condition and an augmented condition. Their mean cross-entropy defines $\mathcal L_{\mathrm{CE}}$, the Jensen--Shannon divergence between their prediction distributions defines $\mathcal L_{\mathrm{JS}}$, and the squared difference between their predicted gate scores defines $\mathcal L_{\mathrm{gate}}$. A task proxy computed from source-training case features also supervises the predicted gate score through mean squared error, denoted $\mathcal L_{\mathrm{proxy}}$. This proxy uses between-class and within-class dispersion of source-case mean patch features and is applied only when sample support is valid. This source-derived proxy regularizes the gate; inference still computes the gate from text alone. The refinement objective is
\begin{align}
 \mathcal L_{\mathrm{refine}}
 &=\mathcal L_{\mathrm{CE}}+0.1\mathcal L_{\mathrm{JS}}\nonumber\\
 &\quad+0.01\mathcal L_{\mathrm{gate}}+0.1\mathcal L_{\mathrm{proxy}}.
 \label{eq:mihyper-refine}
\end{align}
Both stages use source-label supervision; refinement also encourages consistent predictions and gate scores under text augmentation. Each stage runs for 15 epochs. Trainable components use AdamW with a learning rate of $10^{-4}$ and weight decay of $10^{-4}$~\cite{loshchilov2019adamw}, gradient accumulation over 4 steps and a gradient-norm clipping threshold of 1. SOURCE-DEV evaluation is performed every 3 epochs and at the final epoch for checkpoint selection. Algorithm~\ref{alg:mihyper} summarizes training and deployment; evaluation uses the selected checkpoints from the five task folds.

\begin{figure}[t]
\centering
\refstepcounter{algorithm}\label{alg:mihyper}
\definecolor{algcomment}{HTML}{6E9A9B}
\begingroup
\fontsize{8}{9.2}\selectfont\ttfamily
\setlength{\tabcolsep}{0pt}
\renewcommand{\arraystretch}{1}
\begin{tabular*}{\columnwidth}{@{\hspace{5pt}}l@{\extracolsep{\fill}}}
\toprule
\normalfont\fontsize{10}{11.2}\selectfont\textbf{Algorithm \thealgorithm:} MI-Hyper learning and inference\\
\midrule
\textcolor{algcomment}{\# Inputs: source FIT S, DEV D, descriptions,}\\
\textcolor{algcomment}{\# and query bags; encoders remain frozen}\\
\textcolor{algcomment}{\# Sample at most 2048 patches/training bag}\\
\textcolor{algcomment}{\# 1. Joint CE: task context + morphology text}\\
W0, phi = initialize()\\
anchor = fit\_ce(S, W0, phi, text=morphology,\\
\hspace*{7.2em}train=(W0, phi))\\
W0\_star, phi\_star = select\_dev(anchor, D)\\[4pt]
\textcolor{algcomment}{\# 2. Extend generator; fix shared MIL basis}\\
freeze(W0\_star)\\
phi = expand\_dual\_text(phi\_star)\\
objective = CE + 0.1*JS + 0.01*GATE\\
\hspace*{7.2em}+ 0.1*PROXY\\
refined = fit\_full(S, W0\_star, phi,\\
\hspace*{8.4em}loss=objective, train=phi)\\
phi\_hat = select\_dev(refined, D)\\[4pt]
\textcolor{algcomment}{\# 3. Generate from text; infer on all patches}\\
theta = H(phi\_hat, target\_descriptions)\\
probs = [MIL\_all(X, W0\_star, theta)\\
\hspace*{5.4em}for X in target\_bags]\\
\textcolor{algcomment}{\# No target updates; DEV selects checkpoints}\\
\bottomrule
\end{tabular*}
\endgroup
\end{figure}

\subsection{Zero-Shot Inference}

At deployment, the generator and shared basis remain fixed. MI-Hyper generates and caches the aggregation query and classifier parameters once per target text condition, then reuses them across cases. Target labels are not used to generate or update these parameters at deployment. Source-data use is documented separately for the evaluation setting.

The cached parameter count depends on the class vocabulary rather than bag length. Complete bags are processed in chunks with one global attention normalization (Supplementary Section~\ref{sec:si-inference-details}).

Caching avoids regenerating parameters across cases; processing each bag still scales with its patch count.

\FloatBarrier

\definecolor{ExpFocus}{HTML}{F0EAF3}
\definecolor{ExpAux}{HTML}{6D6D6D}
\definecolor{tablepurple}{HTML}{762E80}
\newlength{\ExpContentWidth}
\newcommand{\ExpPair}[2]{\textcolor{ExpAux}{#1}\,/\,#2}
\newcommand{\ExpReadRule}{\specialrule{0.24pt}{1.25pt}{1.25pt}}
\newcommand{\ExpYes}{\ensuremath{\checkmark}}
\newcommand{\ExpNo}{\textcolor{gray}{--}}

\definecolor{GainGreen}{HTML}{008B45}

\definecolor{AllTasksTint}{HTML}{F9E9EF}
\definecolor{HyperTint}{HTML}{EAF3FA}

\section{Experiments}
\label{sec:experiments}

\noindent\textbf{Training details.}
We train five MI-Hyper models under the 112-group partition, using the SOURCE-FIT, SOURCE-DEV and target roles defined in Methods. We optimize parameters on SOURCE-FIT training cases and select checkpoints by task-average AUC on SOURCE-DEV validation cases. Both stages run for 15 epochs with the optimizer and patch-sampling settings in Methods.

\noindent\textbf{Evaluation protocols.}
We evaluate text-conditioned MIL model construction on 308 dataset--endpoint tasks from 52 pathology datasets. The common query set contains 160,989 task--bag records and 896 task-specific classes. Because a case may support several endpoints, these records do not represent mutually independent WSIs or patients. Bags contain 10--145,183 stored patches, with a median of 1,866, and all tasks contribute to the overall statistics. Supplementary Fig.~\ref{fig:registry-overview} summarizes endpoint, anatomical and dataset coverage. Because mutation and staging tasks account for a large share of the collection, we report group-level results alongside the overall averages to distinguish coverage from performance within each group.

For each task, we evaluate its original target-fold checkpoint against MI-Zero on the same query records, using all stored patches and keeping the generator and MIL basis fixed. Seven predefined pairs of text descriptions are used for each task, with a predictor generated and evaluated separately for each condition. Although target task IDs are held out, cases can recur in other source tasks. Matching patient identifiers or feature files across tasks identifies 100,573 query records in the eligible SOURCE-FIT pool and 4,873 in SOURCE-DEV; 55,543 have no recorded match. Pool membership does not establish that every case was sampled during optimization, and an unmatched record does not rule out unrecorded exposure. The evaluation therefore measures transfer to held-out task IDs across the complete query set, rather than generalization to an entirely unseen patient population.

The main text compares macro one-vs-rest AUC and balanced accuracy (BAcc); the complete task-level results additionally report macro-F1 (Supplementary Table~\ref{tab:si-v18-all308}). All metrics are percentages. We first evaluate each condition, average condition metrics within each task, and then weight the 308 tasks equally. MI-Zero uses 50 prompts at each of $K=1,5,10,50,100$, with Mean-$K$ averaging all 250 condition metrics on the same query records as MI-Hyper. MI-Hyper uses all seven predefined text conditions, without choosing conditions or task-specific models from target labels. We compute percentage-point differences from unrounded scores; their signs and colors indicate direction, not statistical significance.

\subsection{Overall Performance of MI-Hyper}
Across all 308 tasks, MI-Hyper achieves mean AUC and BAcc of 58.24 and 45.66, compared with 56.38 and 44.49 for MI-Zero (Mean-$K$), corresponding to gains of $1.85$ and $1.17$ percentage points. Keeping the MI-Hyper configuration fixed, we stratify these results by endpoint type, anatomy, bag structure and dataset source.

Tables~\ref{tab:types} and~\ref{tab:organs} compare MI-Zero and MI-Hyper AUC, BAcc and paired differences across all 16 endpoint types and 25 anatomical display groups, retaining every group and its task count. Supplementary Table~\ref{tab:si-v18-all308} provides all three task-level metrics.

\begin{table}[t]
\captionsetup{type=table}
\centering\begingroup
\definecolor{DeltaGain}{HTML}{008B45}
\definecolor{DeltaDrop}{HTML}{6D6D6D}
\newcommand{\DeltaHead}[1]{{\fontsize{6.6}{8.0}\selectfont #1}}
\fontsize{8.0}{9.8}\selectfont
\setlength{\tabcolsep}{0.65pt}
\setlength{\arrayrulewidth}{0.25pt}
\setlength{\heavyrulewidth}{0.8pt}
\setlength{\lightrulewidth}{0.4pt}
\setlength{\aboverulesep}{0pt}
\setlength{\belowrulesep}{0pt}
\renewcommand{\arraystretch}{1.12}
\setlength{\ExpContentWidth}{\dimexpr\columnwidth-14\tabcolsep-2\arrayrulewidth\relax}
\arrayrulecolor{black}
\begin{tabular*}{\columnwidth}{@{}>{\raggedright\arraybackslash}m{0.215000000000\ExpContentWidth}>{\raggedleft\arraybackslash}m{0.125000000000\ExpContentWidth}|>{\raggedleft\arraybackslash}m{0.101000000000\ExpContentWidth}>{\columncolor{HyperTint}\raggedleft\arraybackslash}m{0.130000000000\ExpContentWidth}>{\raggedleft\arraybackslash}m{0.099000000000\ExpContentWidth}|>{\raggedleft\arraybackslash}m{0.101000000000\ExpContentWidth}>{\columncolor{HyperTint}\raggedleft\arraybackslash}m{0.130000000000\ExpContentWidth}>{\raggedleft\arraybackslash}m{0.099000000000\ExpContentWidth}@{}}
\toprule
\arrayrulecolor{black}
 & & \multicolumn{3}{c|}{\textbf{AUC}$\uparrow$} & \multicolumn{3}{c}{\textbf{BAcc}$\uparrow$} \\
\cline{3-8}
\multirow{-2}{*}{\textbf{\shortstack[l]{Endpoint\\type}}} & \multirow{-2}{*}{\DeltaHead{Task Num}} & \multicolumn{1}{c}{\DeltaHead{MI-Zero}} & \multicolumn{1}{c}{\cellcolor{HyperTint}\DeltaHead{\textbf{MI-Hyper}}} & \multicolumn{1}{c|}{\DeltaHead{$\Delta$ (pp)}} & \multicolumn{1}{c}{\DeltaHead{MI-Zero}} & \multicolumn{1}{c}{\cellcolor{HyperTint}\DeltaHead{\textbf{MI-Hyper}}} & \multicolumn{1}{c}{\DeltaHead{$\Delta$ (pp)}} \\
\midrule
Mutation & 122 & 49.24 & \textbf{54.35} & \textcolor{DeltaGain}{\textbf{+5.11}} & 49.57 & \textbf{53.11} & \textcolor{DeltaGain}{\textbf{+3.54}} \\
Stage & 69 & 54.89 & \textbf{56.85} & \textcolor{DeltaGain}{\textbf{+1.96}} & 38.45 & \textbf{40.71} & \textcolor{DeltaGain}{\textbf{+2.26}} \\
Molecular subtype & 38 & \textbf{55.48} & 54.92 & \textcolor{DeltaDrop}{\textbf{-0.56}} & 30.53 & \textbf{32.31} & \textcolor{DeltaGain}{\textbf{+1.78}} \\
Grade & 29 & 67.65 & \textbf{68.02} & \textcolor{DeltaGain}{\textbf{+0.37}} & 44.35 & \textbf{44.94} & \textcolor{DeltaGain}{\textbf{+0.59}} \\
Histological subtype & 19 & \textbf{81.84} & 72.33 & \textcolor{DeltaDrop}{\textbf{-9.51}} & \textbf{58.94} & 46.37 & \textcolor{DeltaDrop}{\textbf{-12.58}} \\
Immune class & 9 & 51.86 & \textbf{52.36} & \textcolor{DeltaGain}{\textbf{+0.49}} & 33.61 & \textbf{34.06} & \textcolor{DeltaGain}{\textbf{+0.46}} \\
\arrayrulecolor{black}
\specialrule{0.45pt}{0pt}{0pt}
Invasion & 4 & \textbf{79.69} & 76.18 & \textcolor{DeltaDrop}{\textbf{-3.51}} & 55.72 & \textbf{57.65} & \textcolor{DeltaGain}{\textbf{+1.93}} \\
Benign/ malignant & 4 & 73.59 & \textbf{84.49} & \textcolor{DeltaGain}{\textbf{+10.91}} & 50.85 & \textbf{68.55} & \textcolor{DeltaGain}{\textbf{+17.71}} \\
Procedure & 3 & \textbf{63.69} & 56.53 & \textcolor{DeltaDrop}{\textbf{-7.16}} & \textbf{43.27} & 31.13 & \textcolor{DeltaDrop}{\textbf{-12.14}} \\
Tumor detection & 3 & 47.42 & \textbf{51.08} & \textcolor{DeltaGain}{\textbf{+3.66}} & \textbf{49.88} & 43.48 & \textcolor{DeltaDrop}{\textbf{-6.40}} \\
Organ class & 2 & \textbf{96.53} & 72.47 & \textcolor{DeltaDrop}{\textbf{-24.06}} & \textbf{74.68} & 30.66 & \textcolor{DeltaDrop}{\textbf{-44.02}} \\
Receptor status & 2 & \textbf{54.90} & 54.03 & \textcolor{DeltaDrop}{\textbf{-0.86}} & 52.19 & \textbf{52.88} & \textcolor{DeltaGain}{\textbf{+0.69}} \\
\arrayrulecolor{black}
\specialrule{0.45pt}{0pt}{0pt}
Histological pattern & 1 & \textbf{77.29} & 65.97 & \textcolor{DeltaDrop}{\textbf{-11.32}} & \textbf{38.84} & 25.73 & \textcolor{DeltaDrop}{\textbf{-13.11}} \\
Metaplasia & 1 & 63.56 & \textbf{76.37} & \textcolor{DeltaGain}{\textbf{+12.81}} & \textbf{57.83} & 51.58 & \textcolor{DeltaDrop}{\textbf{-6.25}} \\
Metastasis & 1 & 47.17 & \textbf{68.57} & \textcolor{DeltaGain}{\textbf{+21.41}} & 29.40 & \textbf{30.81} & \textcolor{DeltaGain}{\textbf{+1.41}} \\
Molecular status & 1 & 41.97 & \textbf{48.74} & \textcolor{DeltaGain}{\textbf{+6.77}} & 46.46 & \textbf{50.78} & \textcolor{DeltaGain}{\textbf{+4.32}} \\
\arrayrulecolor{black}
\specialrule{0.6pt}{0pt}{0pt}

\rowcolor{AllTasksTint}
\textbf{All tasks} & \textbf{308} & \textbf{56.38} & \textbf{58.24} & \textcolor{DeltaGain}{\textbf{+1.85}} & \textbf{44.49} & \textbf{45.66} & \textcolor{DeltaGain}{\textbf{+1.17}} \\
\arrayrulecolor{black}
\bottomrule
\end{tabular*}
\arrayrulecolor{black}
\endgroup
\captionsetup{position=bottom,skip=4pt}
\caption{\textbf{Results by endpoint type.} AUC and BAcc (\%) for all 16 endpoint types. Task Num denotes the number of tasks; tasks within each group are equally weighted. $\Delta$ is MI-Hyper minus MI-Zero in percentage points, computed before rounding. Gains are green and decreases gray; higher scores in non-summary rows are bold. All three metrics are reported in the supplement.}
\label{tab:types}
\label{tab:group-results}
\end{table}

By endpoint type (Table~\ref{tab:types}), the 122 mutation tasks gain $5.11$ AUC and $3.54$ BAcc points, while the 69 staging tasks gain $1.96$ and $2.26$ points. The four benign/malignant tasks gain $10.91$ and $17.71$ points. For histological subtyping, MI-Hyper achieves 72.33 AUC, compared with 81.84 for MI-Zero. The overall mean therefore combines different endpoint-level results, as the mutation, staging and histological-subtyping comparisons illustrate.

\begin{table}[t]
\captionsetup{type=table}
\centering\begingroup
\definecolor{DeltaGain}{HTML}{008B45}
\definecolor{DeltaDrop}{HTML}{6D6D6D}
\newcommand{\DeltaHead}[1]{{\fontsize{6.6}{8.0}\selectfont #1}}
\fontsize{8.0}{9.8}\selectfont
\setlength{\tabcolsep}{0.65pt}
\setlength{\arrayrulewidth}{0.25pt}
\setlength{\heavyrulewidth}{0.8pt}
\setlength{\lightrulewidth}{0.4pt}
\setlength{\aboverulesep}{0pt}
\setlength{\belowrulesep}{0pt}
\renewcommand{\arraystretch}{1.12}
\setlength{\ExpContentWidth}{\dimexpr\columnwidth-14\tabcolsep-2\arrayrulewidth\relax}
\arrayrulecolor{black}
\begin{tabular*}{\columnwidth}{@{}>{\raggedright\arraybackslash}m{0.215000000000\ExpContentWidth}>{\raggedleft\arraybackslash}m{0.125000000000\ExpContentWidth}|>{\raggedleft\arraybackslash}m{0.101000000000\ExpContentWidth}>{\columncolor{HyperTint}\raggedleft\arraybackslash}m{0.130000000000\ExpContentWidth}>{\raggedleft\arraybackslash}m{0.099000000000\ExpContentWidth}|>{\raggedleft\arraybackslash}m{0.101000000000\ExpContentWidth}>{\columncolor{HyperTint}\raggedleft\arraybackslash}m{0.130000000000\ExpContentWidth}>{\raggedleft\arraybackslash}m{0.099000000000\ExpContentWidth}@{}}
\toprule
\arrayrulecolor{black}
 & & \multicolumn{3}{c|}{\textbf{AUC}$\uparrow$} & \multicolumn{3}{c}{\textbf{BAcc}$\uparrow$} \\
\cline{3-8}
\multirow{-2}{*}{\textbf{Anatomy}} & \multirow{-2}{*}{\DeltaHead{Task Num}} & \multicolumn{1}{c}{\DeltaHead{MI-Zero}} & \multicolumn{1}{c}{\cellcolor{HyperTint}\DeltaHead{\textbf{MI-Hyper}}} & \multicolumn{1}{c|}{\DeltaHead{$\Delta$ (pp)}} & \multicolumn{1}{c}{\DeltaHead{MI-Zero}} & \multicolumn{1}{c}{\cellcolor{HyperTint}\DeltaHead{\textbf{MI-Hyper}}} & \multicolumn{1}{c}{\DeltaHead{$\Delta$ (pp)}} \\
\midrule
Colorectal & 37 & 55.40 & \textbf{59.62} & \textcolor{DeltaGain}{\textbf{+4.22}} & 46.98 & \textbf{50.28} & \textcolor{DeltaGain}{\textbf{+3.30}} \\
Lung & 34 & 55.62 & \textbf{55.87} & \textcolor{DeltaGain}{\textbf{+0.25}} & 44.83 & \textbf{45.50} & \textcolor{DeltaGain}{\textbf{+0.66}} \\
Uterus & 29 & 51.82 & \textbf{60.55} & \textcolor{DeltaGain}{\textbf{+8.74}} & 48.36 & \textbf{54.60} & \textcolor{DeltaGain}{\textbf{+6.24}} \\
Breast & 27 & \textbf{59.28} & 55.66 & \textcolor{DeltaDrop}{\textbf{-3.62}} & \textbf{43.81} & 39.93 & \textcolor{DeltaDrop}{\textbf{-3.88}} \\
Kidney & 24 & 54.91 & \textbf{57.73} & \textcolor{DeltaGain}{\textbf{+2.82}} & 44.03 & \textbf{46.15} & \textcolor{DeltaGain}{\textbf{+2.12}} \\
Stomach & 22 & 59.09 & \textbf{65.77} & \textcolor{DeltaGain}{\textbf{+6.68}} & 47.57 & \textbf{50.57} & \textcolor{DeltaGain}{\textbf{+3.00}} \\
Skin & 19 & 51.10 & \textbf{53.45} & \textcolor{DeltaGain}{\textbf{+2.35}} & 41.47 & \textbf{43.07} & \textcolor{DeltaGain}{\textbf{+1.60}} \\
Bladder & 17 & 59.11 & \textbf{61.81} & \textcolor{DeltaGain}{\textbf{+2.69}} & \textbf{49.42} & 47.54 & \textcolor{DeltaDrop}{\textbf{-1.88}} \\
Brain & 14 & \textbf{52.66} & 44.76 & \textcolor{DeltaDrop}{\textbf{-7.90}} & \textbf{37.60} & 35.63 & \textcolor{DeltaDrop}{\textbf{-1.97}} \\
Prostate & 10 & 58.39 & \textbf{64.21} & \textcolor{DeltaGain}{\textbf{+5.82}} & 36.12 & \textbf{42.27} & \textcolor{DeltaGain}{\textbf{+6.15}} \\
\arrayrulecolor{black}
\specialrule{0.45pt}{0pt}{0pt}
Cervix & 9 & \textbf{66.82} & 63.53 & \textcolor{DeltaDrop}{\textbf{-3.29}} & 47.26 & \textbf{47.43} & \textcolor{DeltaGain}{\textbf{+0.16}} \\
\mbox{Lip (reg.)} & 9 & 51.10 & \textbf{55.15} & \textcolor{DeltaGain}{\textbf{+4.05}} & 31.98 & \textbf{34.50} & \textcolor{DeltaGain}{\textbf{+2.53}} \\
\mbox{Liver/biliary} & 8 & 50.80 & \textbf{53.25} & \textcolor{DeltaGain}{\textbf{+2.45}} & 41.75 & \textbf{42.10} & \textcolor{DeltaGain}{\textbf{+0.35}} \\
Pancreas & 8 & 52.40 & \textbf{57.80} & \textcolor{DeltaGain}{\textbf{+5.40}} & 46.58 & \textbf{51.38} & \textcolor{DeltaGain}{\textbf{+4.81}} \\
Adrenal gland & 7 & \textbf{54.36} & 52.39 & \textcolor{DeltaDrop}{\textbf{-1.97}} & 36.09 & \textbf{37.15} & \textcolor{DeltaGain}{\textbf{+1.06}} \\
Esophagus & 7 & 68.91 & \textbf{74.70} & \textcolor{DeltaGain}{\textbf{+5.79}} & 53.16 & \textbf{61.50} & \textcolor{DeltaGain}{\textbf{+8.34}} \\
Thyroid & 7 & 54.04 & \textbf{54.11} & \textcolor{DeltaGain}{\textbf{+0.07}} & \textbf{35.20} & 34.34 & \textcolor{DeltaDrop}{\textbf{-0.87}} \\
Multiple sites & 4 & \textbf{85.79} & 69.50 & \textcolor{DeltaDrop}{\textbf{-16.29}} & \textbf{58.42} & 30.71 & \textcolor{DeltaDrop}{\textbf{-27.71}} \\
Testis & 4 & \textbf{60.97} & 59.80 & \textcolor{DeltaDrop}{\textbf{-1.17}} & \textbf{50.85} & 49.84 & \textcolor{DeltaDrop}{\textbf{-1.01}} \\
\mbox{Eye/adnexa} & 3 & 50.51 & \textbf{52.66} & \textcolor{DeltaGain}{\textbf{+2.15}} & 44.28 & \textbf{45.25} & \textcolor{DeltaGain}{\textbf{+0.97}} \\
\arrayrulecolor{black}
\specialrule{0.45pt}{0pt}{0pt}
\mbox{Head/neck} & 3 & \textbf{57.94} & 57.57 & \textcolor{DeltaDrop}{\textbf{-0.37}} & 42.50 & \textbf{43.16} & \textcolor{DeltaGain}{\textbf{+0.66}} \\
\mbox{Retro./perit.} & 2 & \textbf{52.72} & 49.46 & \textcolor{DeltaDrop}{\textbf{-3.26}} & \textbf{25.20} & 23.93 & \textcolor{DeltaDrop}{\textbf{-1.27}} \\
Soft tissue & 2 & \textbf{67.84} & 50.14 & \textcolor{DeltaDrop}{\textbf{-17.70}} & \textbf{57.86} & 40.93 & \textcolor{DeltaDrop}{\textbf{-16.93}} \\
Lymph node & 1 & 47.17 & \textbf{68.57} & \textcolor{DeltaGain}{\textbf{+21.41}} & 29.40 & \textbf{30.81} & \textcolor{DeltaGain}{\textbf{+1.41}} \\
Thymus & 1 & \textbf{77.96} & 64.23 & \textcolor{DeltaDrop}{\textbf{-13.73}} & 51.22 & \textbf{60.44} & \textcolor{DeltaGain}{\textbf{+9.22}} \\
\arrayrulecolor{black}
\specialrule{0.6pt}{0pt}{0pt}

\rowcolor{AllTasksTint}
\textbf{All tasks} & \textbf{308} & \textbf{56.38} & \textbf{58.24} & \textcolor{DeltaGain}{\textbf{+1.85}} & \textbf{44.49} & \textbf{45.66} & \textcolor{DeltaGain}{\textbf{+1.17}} \\
\arrayrulecolor{black}
\bottomrule
\end{tabular*}
\arrayrulecolor{black}
\endgroup
\captionsetup{position=bottom,skip=4pt}
\caption{\textbf{Results by anatomical group.} AUC and BAcc (\%) for all 25 display groups. Task Num denotes the number of tasks; tasks within each group are equally weighted. Formatting follows Table~\ref{tab:types}. Lip (reg.) retains the registry label; Retro./perit. denotes retroperitoneum/peritoneum. Full group names and all three metrics are given in the supplement.}
\label{tab:organs}
\end{table}

By anatomy (Table~\ref{tab:organs}), the 37 colorectal tasks gain $4.22$ AUC and $3.30$ BAcc points, and the 29 uterine tasks gain $8.74$ and $6.24$ points. Anatomical groups differ in endpoint composition, so their means reflect both anatomical coverage and task composition. Supplementary Fig.~\ref{fig:registry-overview} shows uneven task coverage across organs. These registry groupings do not evaluate transfer to held-out organs.

Table~\ref{tab:bag-structure} connects performance to sequence length and class count. On the 53 tasks whose average bag length exceeds 5,000 patches, MI-Hyper gains $7.54$ AUC and $5.51$ BAcc points. Inference uses all stored patches, whereas training follows the sampling strategy in Methods. Because endpoints, sources and class counts also vary across length groups, these comparisons show performance on long bags without isolating a causal effect of sequence length.

\begin{table}[t]
\captionsetup{type=table}
\centering\begingroup
\definecolor{DeltaGain}{HTML}{008B45}
\definecolor{DeltaDrop}{HTML}{6D6D6D}
\newcommand{\DeltaHead}[1]{{\fontsize{6.3}{7.6}\selectfont #1}}
\newcommand{\DeltaSide}[1]{{\fontsize{7.2}{8.6}\selectfont #1}}
\fontsize{8.0}{9.8}\selectfont
\setlength{\tabcolsep}{0.65pt}
\setlength{\arrayrulewidth}{0.25pt}
\setlength{\heavyrulewidth}{0.8pt}
\setlength{\lightrulewidth}{0.4pt}
\setlength{\aboverulesep}{0pt}
\setlength{\belowrulesep}{0pt}
\renewcommand{\arraystretch}{1.12}
\setlength{\ExpContentWidth}{\dimexpr\columnwidth-16\tabcolsep-3\arrayrulewidth\relax}
\arrayrulecolor{black}
\begin{tabular*}{\columnwidth}{@{}>{\centering\arraybackslash}m{0.105000000000\ExpContentWidth}|>{\raggedright\arraybackslash}m{0.205000000000\ExpContentWidth}>{\raggedleft\arraybackslash}m{0.074000000000\ExpContentWidth}|>{\raggedleft\arraybackslash}m{0.098000000000\ExpContentWidth}>{\columncolor{HyperTint}\raggedleft\arraybackslash}m{0.126000000000\ExpContentWidth}>{\raggedleft\arraybackslash}m{0.084000000000\ExpContentWidth}|>{\raggedleft\arraybackslash}m{0.098000000000\ExpContentWidth}>{\columncolor{HyperTint}\raggedleft\arraybackslash}m{0.126000000000\ExpContentWidth}>{\raggedleft\arraybackslash}m{0.084000000000\ExpContentWidth}@{}}
\toprule

 & & & \multicolumn{3}{c|}{\textbf{AUC}$\uparrow$} & \multicolumn{3}{c}{\textbf{BAcc}$\uparrow$} \\
\cline{4-9}

\multirow{-2}{*}{\DeltaHead{Axis}} & \multirow{-2}{*}{\DeltaHead{\textbf{Group}}} & \multirow{-2}{*}{\DeltaHead{\shortstack{Task\\Num}}} & \multicolumn{1}{c}{\DeltaHead{MI-Zero}} & \multicolumn{1}{c}{\cellcolor{HyperTint}\DeltaHead{\textbf{MI-Hyper}}} & \multicolumn{1}{c|}{\DeltaHead{$\Delta$ (pp)}} & \multicolumn{1}{c}{\DeltaHead{MI-Zero}} & \multicolumn{1}{c}{\cellcolor{HyperTint}\DeltaHead{\textbf{MI-Hyper}}} & \multicolumn{1}{c}{\DeltaHead{$\Delta$ (pp)}} \\
\midrule
\multirow{4}{*}{\DeltaSide{\shortstack{Bag\\length}}} & $\leq1{,}000$ & 33 & \textbf{67.04} & 66.90 & \textcolor{DeltaDrop}{\textbf{-0.15}} & \textbf{49.00} & 45.55 & \textcolor{DeltaDrop}{\textbf{-3.44}} \\
 & \mbox{$1{,}000$--$2{,}500$} & 27 & \textbf{59.46} & 57.69 & \textcolor{DeltaDrop}{\textbf{-1.77}} & \textbf{47.93} & 47.29 & \textcolor{DeltaDrop}{\textbf{-0.63}} \\
 & \mbox{$2{,}500$--$5{,}000$} & 195 & 55.47 & \textbf{56.62} & \textcolor{DeltaGain}{\textbf{+1.15}} & 42.49 & \textbf{43.51} & \textcolor{DeltaGain}{\textbf{+1.02}} \\
 & $>5{,}000$ & 53 & 51.52 & \textbf{59.06} & \textcolor{DeltaGain}{\textbf{+7.54}} & 47.27 & \textbf{52.78} & \textcolor{DeltaGain}{\textbf{+5.51}} \\
\midrule
\multirow{2}{*}{\DeltaSide{\shortstack{Class\\count}}} & Binary & 174 & 53.51 & \textbf{57.37} & \textcolor{DeltaGain}{\textbf{+3.86}} & 51.83 & \textbf{54.57} & \textcolor{DeltaGain}{\textbf{+2.73}} \\
 & Multiclass & 134 & \textbf{60.11} & 59.36 & \textcolor{DeltaDrop}{\textbf{-0.75}} & \textbf{34.95} & 34.08 & \textcolor{DeltaDrop}{\textbf{-0.86}} \\
\bottomrule
\end{tabular*}
\arrayrulecolor{black}
\endgroup
\captionsetup{position=bottom,skip=4pt}
\caption{\textbf{Results by bag length and class count.} Task Num denotes the number of tasks. Length groups use the task-average patch count $\bar N$: $\bar N\leq1{,}000$, $1{,}000<\bar N\leq2{,}500$, $2{,}500<\bar N\leq5{,}000$ and $\bar N>5{,}000$. Class-count groups distinguish binary and multiclass tasks. Each partition covers all 308 tasks. AUC and BAcc are percentages. Formatting follows Table~\ref{tab:types}. These groups do not isolate length robustness.}
\label{tab:bag-structure}
\end{table}

Grouping by dataset source gives a similar comparison (Table~\ref{tab:sources-all}). Across the 223 TCGA tasks, MI-Hyper achieves 56.68 AUC and 45.92 BAcc, gaining $2.98$ and $2.87$ points over MI-Zero and contributing substantially to the overall gains. The 52 additional-source tasks change by $-0.41$ and $-4.11$ points, while the 33 CPTAC tasks change by $-2.20$ and $-2.02$ points. Supplementary Fig.~\ref{fig:dataset-task-catalogue} links these source groups to the endpoints represented in each dataset. Source averages should be interpreted alongside endpoint composition. All groups use the common query set and established checkpoints, so these comparisons do not evaluate transfer to held-out dataset sources.

\begin{table}[t]
\captionsetup{type=table}
\centering\begingroup
\definecolor{DeltaGain}{HTML}{008B45}
\definecolor{DeltaDrop}{HTML}{6D6D6D}
\newcommand{\DeltaHead}[1]{{\fontsize{6.6}{8.0}\selectfont #1}}
\fontsize{8.0}{9.8}\selectfont
\setlength{\tabcolsep}{0.65pt}
\setlength{\arrayrulewidth}{0.25pt}
\setlength{\heavyrulewidth}{0.8pt}
\setlength{\lightrulewidth}{0.4pt}
\setlength{\aboverulesep}{0pt}
\setlength{\belowrulesep}{0pt}
\renewcommand{\arraystretch}{1.12}
\setlength{\ExpContentWidth}{\dimexpr\columnwidth-14\tabcolsep-2\arrayrulewidth\relax}
\arrayrulecolor{black}
\begin{tabular*}{\columnwidth}{@{}>{\raggedright\arraybackslash}m{0.195000000000\ExpContentWidth}>{\raggedleft\arraybackslash}m{0.125000000000\ExpContentWidth}|>{\raggedleft\arraybackslash}m{0.106000000000\ExpContentWidth}>{\columncolor{HyperTint}\raggedleft\arraybackslash}m{0.130000000000\ExpContentWidth}>{\raggedleft\arraybackslash}m{0.104000000000\ExpContentWidth}|>{\raggedleft\arraybackslash}m{0.106000000000\ExpContentWidth}>{\columncolor{HyperTint}\raggedleft\arraybackslash}m{0.130000000000\ExpContentWidth}>{\raggedleft\arraybackslash}m{0.104000000000\ExpContentWidth}@{}}
\toprule
\arrayrulecolor{black}
 & & \multicolumn{3}{c|}{\textbf{AUC}$\uparrow$} & \multicolumn{3}{c}{\textbf{BAcc}$\uparrow$} \\
\cline{3-8}
\multirow{-2}{*}{\textbf{Source}} & \multirow{-2}{*}{\DeltaHead{Task Num}} & \multicolumn{1}{c}{\DeltaHead{MI-Zero}} & \multicolumn{1}{c}{\cellcolor{HyperTint}\DeltaHead{\textbf{MI-Hyper}}} & \multicolumn{1}{c|}{\DeltaHead{$\Delta$ (pp)}} & \multicolumn{1}{c}{\DeltaHead{MI-Zero}} & \multicolumn{1}{c}{\cellcolor{HyperTint}\DeltaHead{\textbf{MI-Hyper}}} & \multicolumn{1}{c}{\DeltaHead{$\Delta$ (pp)}} \\
\midrule
TCGA & 223 & 53.70 & \textbf{56.68} & \textcolor{DeltaGain}{\textbf{+2.98}} & 43.05 & \textbf{45.92} & \textcolor{DeltaGain}{\textbf{+2.87}} \\
Additional & 52 & \textbf{69.18} & 68.76 & \textcolor{DeltaDrop}{\textbf{-0.41}} & \textbf{48.67} & 44.56 & \textcolor{DeltaDrop}{\textbf{-4.11}} \\
CPTAC & 33 & \textbf{54.33} & 52.13 & \textcolor{DeltaDrop}{\textbf{-2.20}} & \textbf{47.59} & 45.56 & \textcolor{DeltaDrop}{\textbf{-2.02}} \\
\specialrule{0.6pt}{0pt}{0pt}

\rowcolor{AllTasksTint}
\textbf{All tasks} & \textbf{308} & \textbf{56.38} & \textbf{58.24} & \textcolor{DeltaGain}{\textbf{+1.85}} & \textbf{44.49} & \textbf{45.66} & \textcolor{DeltaGain}{\textbf{+1.17}} \\
\bottomrule
\end{tabular*}
\arrayrulecolor{black}
\endgroup
\captionsetup{position=bottom,skip=4pt}
\caption{\textbf{Results by dataset source.} AUC and BAcc (\%) with Task Num denoting the number of tasks. All 308 tasks are retained and equally weighted within each source. Source grouping does not alter the five-fold partition of 112 semantic groups. Formatting follows Table~\ref{tab:types}.}
\label{tab:sources-all}
\end{table}

Beyond the aggregate gains, MI-Hyper produces larger improvements on some individual tasks. Table~\ref{tab:concurrent-gains} reports AUC, BAcc and paired differences for six tasks spanning benign/malignant classification, molecular subtyping, mutation prediction and tumor grading. The display rule starts from all 308 tasks and requires at least 100 bags, at least 20 bags per class, and gains of at least ten percentage points in each of the three metrics reported in Supplementary Table~\ref{tab:si-v18-all308}. We rank eligible tasks by their smallest gain across the three metrics, retain the highest-ranked task in each anatomical group and show the first six. This display selection leaves the complete task set and overall scores unchanged.

\begin{table}[t]
\captionsetup{type=table}
\centering\begingroup
\definecolor{DeltaGain}{HTML}{008B45}
\definecolor{DeltaDrop}{HTML}{6D6D6D}
\newcommand{\DeltaHead}[1]{{\fontsize{6.6}{8.0}\selectfont #1}}
\fontsize{8.0}{9.8}\selectfont
\setlength{\tabcolsep}{0.65pt}
\setlength{\arrayrulewidth}{0.25pt}
\setlength{\heavyrulewidth}{0.8pt}
\setlength{\lightrulewidth}{0.4pt}
\setlength{\aboverulesep}{0pt}
\setlength{\belowrulesep}{0pt}
\renewcommand{\arraystretch}{1.12}
\setlength{\ExpContentWidth}{\dimexpr\columnwidth-14\tabcolsep-2\arrayrulewidth\relax}
\arrayrulecolor{black}
\begin{tabular*}{\columnwidth}{@{}>{\raggedright\arraybackslash}m{0.265000000000\ExpContentWidth}>{\raggedleft\arraybackslash}m{0.075000000000\ExpContentWidth}|>{\raggedleft\arraybackslash}m{0.102000000000\ExpContentWidth}>{\columncolor{HyperTint}\raggedleft\arraybackslash}m{0.130000000000\ExpContentWidth}>{\raggedleft\arraybackslash}m{0.098000000000\ExpContentWidth}|>{\raggedleft\arraybackslash}m{0.102000000000\ExpContentWidth}>{\columncolor{HyperTint}\raggedleft\arraybackslash}m{0.130000000000\ExpContentWidth}>{\raggedleft\arraybackslash}m{0.098000000000\ExpContentWidth}@{}}
\toprule
\arrayrulecolor{black}

 & & \multicolumn{3}{c|}{\textbf{AUC}$\uparrow$} & \multicolumn{3}{c}{\textbf{BAcc}$\uparrow$} \\
\cline{3-8}

\multirow{-2}{*}{\textbf{Task}} & \multirow{-2}{*}{$n$} & \multicolumn{1}{c}{\DeltaHead{MI-Zero}} & \multicolumn{1}{c}{\cellcolor{HyperTint}\DeltaHead{\textbf{MI-Hyper}}} & \multicolumn{1}{c|}{\DeltaHead{$\Delta$ (pp)}} & \multicolumn{1}{c}{\DeltaHead{MI-Zero}} & \multicolumn{1}{c}{\cellcolor{HyperTint}\DeltaHead{\textbf{MI-Hyper}}} & \multicolumn{1}{c}{\DeltaHead{$\Delta$ (pp)}} \\
\midrule
\shortstack[l]{Lung benign/\\malignant} & 888 & 69.42 & \textbf{94.43} & \textcolor{DeltaGain}{+25.01} & 65.58 & \textbf{86.67} & \textcolor{DeltaGain}{+21.09} \\
\specialrule{0.2pt}{1.1pt}{1.1pt}
\shortstack[l]{TCGA-ESCA\\subtype\\(CIN vs.\ ESCC)} & 141 & 76.96 & \textbf{94.63} & \textcolor{DeltaGain}{+17.67} & 62.15 & \textbf{84.25} & \textcolor{DeltaGain}{+22.09} \\
\specialrule{0.2pt}{1.1pt}{1.1pt}
\shortstack[l]{COAD ATRX\\mutation} & 434 & 42.35 & \textbf{65.81} & \textcolor{DeltaGain}{+23.46} & 45.96 & \textbf{62.86} & \textcolor{DeltaGain}{+16.91} \\
\specialrule{0.2pt}{1.1pt}{1.1pt}
\shortstack[l]{UCEC SETD2\\mutation} & 505 & 46.47 & \textbf{71.46} & \textcolor{DeltaGain}{+25.00} & 47.92 & \textbf{64.52} & \textcolor{DeltaGain}{+16.60} \\
\specialrule{0.2pt}{1.1pt}{1.1pt}
\shortstack[l]{SKCM mRNA\\subtype} & 291 & 51.36 & \textbf{65.59} & \textcolor{DeltaGain}{+14.23} & 29.69 & \textbf{45.04} & \textcolor{DeltaGain}{+15.34} \\
\specialrule{0.2pt}{1.1pt}{1.1pt}
HNSC grade & 428 & 54.19 & \textbf{67.22} & \textcolor{DeltaGain}{+13.03} & 32.91 & \textbf{48.51} & \textcolor{DeltaGain}{+15.60} \\
\bottomrule
\end{tabular*}
\arrayrulecolor{black}
\endgroup
\captionsetup{position=bottom,skip=4pt}
\caption{\textbf{Six tasks with substantial gains.} AUC and BAcc are percentages; $n$ is the number of bags. Higher scores are bold. $\Delta$ is MI-Hyper minus MI-Zero in percentage points, computed before rounding, with gains in green. The lung task is from REG\_dataset; ESCA uses the source-defined labels CIN and ESCC. Each example has at least 100 bags and 20 bags per class; the selection rule is given in the text. All 308 tasks contribute to the overall results.}
\label{tab:concurrent-gains}
\end{table}

Lung benign/malignant classification reaches 94.43 AUC and 86.67 BAcc on 888 bags, gaining 25.01 and 21.09 points; its smallest class contains 126 bags. The source-defined TCGA-ESCA subtype task (CIN vs.\ ESCC) reaches 94.63 AUC and 84.25 BAcc on 141 bags, gaining 17.67 and 22.09 points. Both tasks improve in AUC and BAcc, reflecting concurrent gains in ranking and class decisions.

Molecular-marker tasks provide complementary examples. COAD ATRX AUC increases from 42.35 to 65.81, with a 16.91-point BAcc gain, while UCEC SETD2 gains 25.00 AUC and 16.60 BAcc points. These tasks contain 434 and 505 bags, with smallest classes of 32 and 62 bags. SKCM mRNA subtyping reaches 65.59 AUC and 45.04 BAcc, and HNSC grading reaches 67.22 AUC and 48.51 BAcc. Because starting scores and final performance differ across endpoints, the table retains absolute scores, paired changes and sample counts to place each improvement in context.

Across the complete collection, MI-Hyper improves AUC on 182 tasks and BAcc on 184 tasks. The median AUC change is $+1.94$ points, with tenth and ninetieth percentiles of $-12.97$ and $+17.02$. Supplementary Table~\ref{tab:si-v18-all308} reports every task, allowing these larger gains to be viewed within the full result distribution.

\subsection{Generated Queries and Spatial Attention}
\begin{figure}[!t]
\centering
\includegraphics[width=\linewidth]{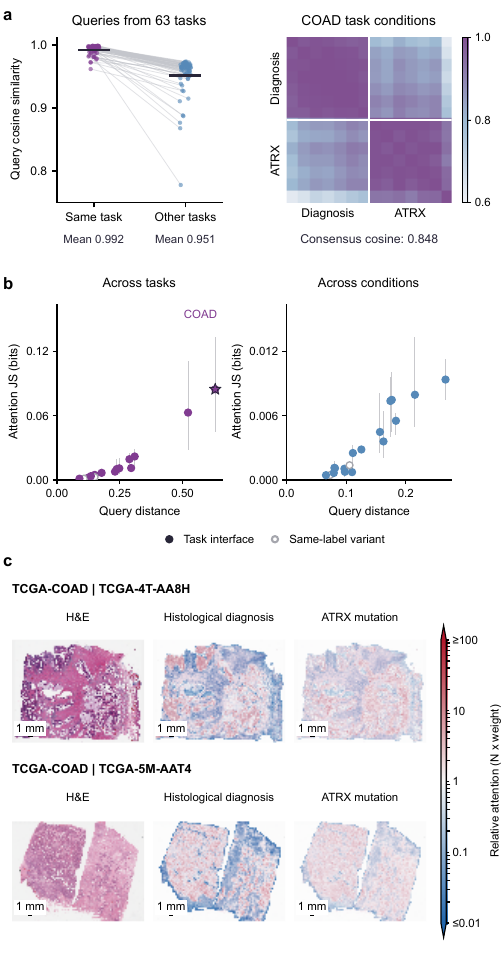}
\caption{Query similarities and attention maps. (a) Within- and between-task query cosine similarities for 63 tasks and the condition-level similarity matrix for the COAD diagnosis and ATRX mutation tasks. (b) Query distance and Jensen--Shannon divergence of attention distributions on 18 WSIs. Bars show the range across three cases; open markers denote same-label variants. (c) Two COAD WSIs, each shown with H\&E and attention maps for histological diagnosis and ATRX mutation prediction. Each map is the mean of seven separately normalized attention distributions. Colors show $N\bar{\alpha}$ (0.01--100): blue indicates lower and red higher relative attention; 1 denotes uniform attention. Scale bars: 1\,mm.}
\label{fig:weight-attention}
\label{fig:query-similarity}
\label{fig:tcga-attention}
\end{figure}

Figure~\ref{fig:weight-attention}a includes all 63 target tasks associated with the checkpoint used for the COAD examples. Within each task, we average the 21 pairwise cosine similarities among its seven condition-specific queries, yielding a task-average value of 0.992. For between-task comparisons, we average each task's seven unit-normalized queries and normalize the resulting consensus. Each task is compared with the other 62 task consensuses; the task-average cosine similarity is 0.951. The consensus cosine between COAD histological diagnosis and ATRX mutation prediction is 0.848.

We next relate query differences to attention differences on the same patch sets across 18 WSIs from six TCGA datasets (Fig.~\ref{fig:weight-attention}b). Query distance is the Euclidean distance divided by the mean query norm, $d(q,r)=2\lVert q-r\rVert_2/(\lVert q\rVert_2+\lVert r\rVert_2)$. Attention differences use base-2 Jensen--Shannon (JS) divergence. For each WSI, we average 49 cross-task or 21 within-task condition pairs, then average across three slides for each task or task pair. Spearman correlations across task-pair means and task means are 0.959 and 0.940, respectively. In these cases, query distance is positively associated with attention divergence.

Pooling depends on query direction and magnitude, but cosine similarity captures only direction. Positive rescaling can therefore change attention concentration without changing query direction. For two patches in the same bag, Eq.~\eqref{eq:pooling} gives $\log(\alpha_i/\alpha_j)=q_\tau^\top(g_i-g_j)$. Consequently, a query change can alter relative patch weights through its projection onto the patch-feature difference, even when query directions remain close. Figure~\ref{fig:weight-attention}b therefore compares generated queries with the attention distributions they induce on fixed patch sets.

On each COAD WSI, histological diagnosis and ATRX mutation prediction use the same patches, generator and shared MIL basis (Fig.~\ref{fig:weight-attention}c). The difference in relative tissue weights therefore arises from the generated task query. Supplementary Fig.~\ref{fig:si-task-attention} shows 16 distinct tasks and WSIs from six TCGA datasets, covering grading, molecular subtyping and mutation prediction. Unlike the COAD comparisons, each supplementary panel uses a different task and WSI.

\begin{figure}[!t]
\centering
\includegraphics[width=\columnwidth]{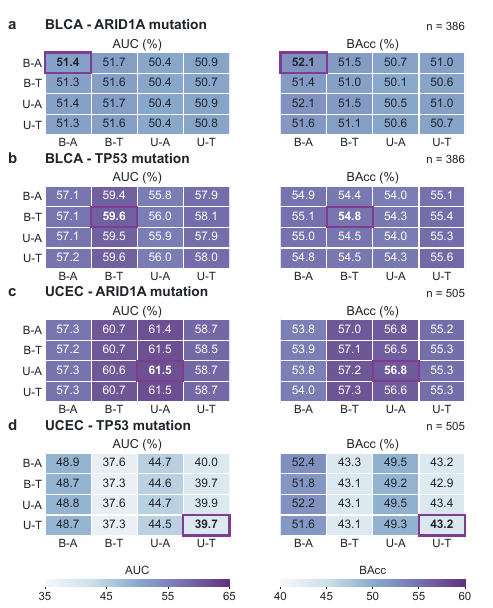}
\caption{Query--classifier recombination. Rows denote query donors and columns classifier $(W,b)$ donors. B-A, B-T, U-A and U-T denote BLCA ARID1A, BLCA TP53, UCEC ARID1A and UCEC TP53. Cells give AUC or BAcc (\%) averaged over seven conditions on complete target case pools. Purple outlines mark original pairs. Across genes, class rows align wildtype and mutant states.}
\label{fig:parameter-exchange}
\end{figure}

\subsection{Query--Classifier Recombination}
Having related generated queries to spatial attention, we next examine how the query and classifier affect case-level ranking. We exchange their sources across four tasks: ARID1A and TP53 prediction in BLCA and UCEC (Fig.~\ref{fig:parameter-exchange}). All combinations use the same frozen MIL basis and target-case features. Classifier rows are aligned by wildtype and mutant labels before exchange. We evaluate 891 cases and average metrics over seven separately evaluated text conditions.

For each target, we first keep its original classifier fixed and compare the AUC obtained with queries from all four tasks. The difference between the highest and lowest AUC averages 0.19 percentage points over the four targets. We then keep its original query fixed and exchange classifier weights and biases. This gives an average AUC range of 5.13 percentage points. Therefore, ranking performance is more sensitive to classifier source than to query source in these tasks.

\section{Conclusion}
Through mixed-task source training, MI-Hyper learns to generate the aggregation and classification weights of MIL predictors from pathology task descriptions. Across 308 tasks from 52 pathology datasets, its mean AUC and balanced accuracy exceed those of MI-Zero, with substantial concurrent gains on some endpoints. These findings provide experimental support for constructing WSI predictors from task descriptions. Generating one predictor per text condition and reusing it across cases offers a promising route to weight-space transfer for long visual sequences.

Future work should broaden source-task coverage and investigate data augmentation. More stable weight generation will also be important for reliable prediction.

\label{page:main-end}

\clearpage
\section*{LLM Usage Statement}
\label{sec:llm-usage}

We used OpenAI Codex to revise the manuscript, check experimental settings and result records, and prepare tables and figures from saved measurements. All reported performance values come from existing evaluation files; none are simulated. The source package includes results and verification scripts. Schematics use real WSI thumbnails and matching crops; prediction bars and model components are illustrative. Displayed task examples are subsets of the complete results, and parameter counts and computational analyses are not runtime measurements. The authors are responsible for verifying the scientific content, citations and final submission.

{\small

\bibliographystyle{ieeenat_fullname}

\bibliography{references_methods9_20261116}

@inproceedings{ilse2018attention,
  title = {Attention-based Deep Multiple Instance Learning},
  author = {Ilse, Maximilian and Tomczak, Jakub and Welling, Max},
  booktitle = {Proceedings of the 35th International Conference on Machine Learning},
  series = {Proceedings of Machine Learning Research},
  volume = {80},
  pages = {2127--2136},
  year = {2018},
  publisher = {PMLR},
  url = {https://proceedings.mlr.press/v80/ilse18a.html}
}

@article{lu2021clam,
  title = {Data-efficient and weakly supervised computational pathology on whole-slide images},
  author = {Lu, Ming Y. and Williamson, Drew F. K. and Chen, Tiffany Y. and Chen, Richard J. and Barbieri, Matteo and Mahmood, Faisal},
  journal = {Nature Biomedical Engineering},
  volume = {5},
  pages = {555--570},
  year = {2021},
  doi = {10.1038/s41551-020-00682-w},
  url = {https://www.nature.com/articles/s41551-020-00682-w}
}

@inproceedings{shao2021transmil,
  title = {{TransMIL}: Transformer based Correlated Multiple Instance Learning for Whole Slide Image Classification},
  author = {Shao, Zhuchen and Bian, Hao and Chen, Yang and Wang, Yifeng and Zhang, Jian and Ji, Xiangyang and Zhang, Yongbing},
  booktitle = {Advances in Neural Information Processing Systems},
  volume = {34},
  pages = {2136--2147},
  year = {2021},
  url = {https://proceedings.neurips.cc/paper/2021/hash/10c272d06794d3e5785d5e7c5356e9ff-Abstract.html}
}

@inproceedings{lu2023mizero,
  title = {Visual Language Pretrained Multiple Instance Zero-Shot Transfer for Histopathology Images},
  author = {Lu, Ming Y. and Chen, Bowen and Zhang, Andrew and Williamson, Drew F. K. and Chen, Richard J. and Ding, Tong and Le, Long Phi and Chuang, Yung-Sung and Mahmood, Faisal},
  booktitle = {Proceedings of the IEEE/CVF Conference on Computer Vision and Pattern Recognition},
  pages = {19764--19775},
  year = {2023},
  url = {https://openaccess.thecvf.com/content/CVPR2023/html/Lu_Visual_Language_Pretrained_Multiple_Instance_Zero-Shot_Transfer_for_Histopathology_Images_CVPR_2023_paper.html}
}

@article{lu2024conch,
  title = {A visual-language foundation model for computational pathology},
  author = {Lu, Ming Y. and Chen, Bowen and Williamson, Drew F. K. and Chen, Richard J. and Liang, Ivy and Ding, Tong and Jaume, Guillaume and Odintsov, Igor and Le, Long Phi and Gerber, Georg and Parwani, Anil V. and Zhang, Andrew and Mahmood, Faisal},
  journal = {Nature Medicine},
  volume = {30},
  pages = {863--874},
  year = {2024},
  doi = {10.1038/s41591-024-02856-4},
  url = {https://www.nature.com/articles/s41591-024-02856-4}
}

@article{ding2025titan,
  title = {A multimodal whole-slide foundation model for pathology},
  author = {Ding, Tong and Wagner, Sophia J. and Song, Andrew H. and Chen, Richard J. and Lu, Ming Y. and Zhang, Andrew and Vaidya, Anurag J. and Jaume, Guillaume and Shaban, Muhammad and Kim, Ahrong and Williamson, Drew F. K. and Robertson, Harry and Chen, Bowen and Almagro-P{\'e}rez, Cristina and Doucet, Paul and Sahai, Sharifa and Chen, Chengkuan and Chen, Christina S. and Komura, Daisuke and Kawabe, Akihiro and Ochi, Mieko and Sato, Shinya and Yokose, Tomoyuki and Miyagi, Yohei and Ishikawa, Shumpei and Gerber, Georg and Peng, Tingying and Le, Long Phi and Mahmood, Faisal},
  journal = {Nature Medicine},
  volume = {31},
  pages = {3749--3761},
  year = {2025},
  doi = {10.1038/s41591-025-03982-3},
  url = {https://www.nature.com/articles/s41591-025-03982-3}
}

@article{shaikovski2024prism,
  title = {{PRISM}: A Multi-Modal Generative Foundation Model for Slide-Level Histopathology},
  author = {Shaikovski, George and Casson, Adam and Severson, Kristen and Zimmermann, Eric and Wang, Yi Kan and Kunz, Jeremy D. and Retamero, Juan A. and Oakley, Gerard and Klimstra, David and Kanan, Christopher and Hanna, Matthew and Zelechowski, Michal and Viret, Julian and Tenenholtz, Neil and Hall, James and Fusi, Nicolo and Yousfi, Razik and Hamilton, Peter and Moye, William A. and Vorontsov, Eugene and Liu, Siqi and Fuchs, Thomas J.},
  journal = {arXiv preprint arXiv:2405.10254},
  year = {2024},
  url = {https://arxiv.org/abs/2405.10254}
}

@inproceedings{ha2017hypernetworks,
  title = {{HyperNetworks}},
  author = {Ha, David and Dai, Andrew and Le, Quoc V.},
  booktitle = {International Conference on Learning Representations},
  year = {2017},
  url = {https://openreview.net/forum?id=rkpACe1lx}
}

@inproceedings{hu2022lora,
  title = {{LoRA}: Low-Rank Adaptation of Large Language Models},
  author = {Hu, Edward J. and Shen, Yelong and Wallis, Phillip and Allen-Zhu, Zeyuan and Li, Yuanzhi and Wang, Shean and Wang, Lu and Chen, Weizhu},
  booktitle = {International Conference on Learning Representations},
  year = {2022},
  url = {https://openreview.net/forum?id=nZeVKeeFYf9}
}

@inproceedings{amosy2024text2model,
  title = {{Text2Model}: Text-based Model Induction for Zero-shot Image Classification},
  author = {Amosy, Ohad and Volk, Tomer and Shapira, Eilam and Ben-David, Eyal and Reichart, Roi and Chechik, Gal},
  booktitle = {Findings of the Association for Computational Linguistics: EMNLP 2024},
  pages = {155--172},
  year = {2024},
  publisher = {Association for Computational Linguistics},
  doi = {10.18653/v1/2024.findings-emnlp.8},
  url = {https://aclanthology.org/2024.findings-emnlp.8/}
}

@inproceedings{chen2025texttolora,
  title = {Text-to-{LoRA}: Instant Transformer Adaption},
  author = {Charakorn, Rujikorn and Cetin, Edoardo and Tang, Yujin and Lange, Robert Tjarko},
  booktitle = {Proceedings of the 42nd International Conference on Machine Learning},
  series = {Proceedings of Machine Learning Research},
  volume = {267},
  pages = {7485--7514},
  year = {2025},
  publisher = {PMLR},
  url = {https://proceedings.mlr.press/v267/charakorn25a.html}
}

@inproceedings{radford2021clip,
  title = {Learning Transferable Visual Models From Natural Language Supervision},
  author = {Radford, Alec and Kim, Jong Wook and Hallacy, Chris and Ramesh, Aditya and Goh, Gabriel and Agarwal, Sandhini and Sastry, Girish and Askell, Amanda and Mishkin, Pamela and Clark, Jack and Krueger, Gretchen and Sutskever, Ilya},
  booktitle = {Proceedings of the 38th International Conference on Machine Learning},
  series = {Proceedings of Machine Learning Research},
  volume = {139},
  pages = {8748--8763},
  year = {2021},
  publisher = {PMLR},
  url = {https://proceedings.mlr.press/v139/radford21a.html}
}

@inproceedings{shao2026limit,
  title = {Beyond Sparse Captions: Aligning Slide-Level Text and Patch-Level Vision in Pathology},
  author = {Shao, Daniel and Bang, Dongmin and Weishaupt, Luca L. and Wagner, Sophia J. and Reitsam, Nic G. and Lu, Ming Y. and Le, Long P. and Chen, Richard J. and Mahmood, Faisal},
  booktitle = {Advances in Neural Information Processing Systems},
  year = {2026},
  url = {https://nips.cc/virtual/2026/poster/149751}
}

@inproceedings{zhang2026care,
  title = {{CARE}: A Molecular-Guided Foundation Model with Adaptive Region Modeling for Whole Slide Image Analysis},
  author = {Zhang, Di and Gong, Zhangpeng and Pang, Xiaobo and Liu, Jiashuai and Lu, Junbo and Cui, Hao and Ge, Jiusong and Zeng, Zhi and Yi, Kai and Li, Yinghua and Liu, Si and Yu, Tingsong and Wang, Haoran and Crispin-Ortuzar, Mireia and Yu, Weimiao and Li, Chen and Gao, Zeyu},
  booktitle = {Proceedings of the IEEE/CVF Conference on Computer Vision and Pattern Recognition},
  year = {2026},
  pages = {21078--21088},
  url = {https://openaccess.thecvf.com/content/CVPR2026/html/Zhang_CARE_A_Molecular-Guided_Foundation_Model_with_Adaptive_Region_Modeling_for_CVPR_2026_paper.html}
}

@article{vorontsov2026prism2,
  title = {End-to-end multimodal pathology foundation model with clinical dialogue},
  author = {Vorontsov, Eugene and Shaikovski, George and Casson, Adam and Viret, Julian and Zimmermann, Eric and Tenenholtz, Neil and Wang, Yi Kan and Bernhard, Jan H. and Godrich, Ran A. and Retamero, Juan A. and Shia, Jinru and Gonen, Mithat and Weiser, Martin R. and Klimstra, David S. and Yousfi, Razik and Fusi, Nicol{\`o} and Fuchs, Thomas J. and Severson, Kristen and Liu, Siqi},
  journal = {Nature Medicine},
  volume = {32},
  pages = {3203--3213},
  year = {2026},
  doi = {10.1038/s41591-026-04521-4},
  url = {https://www.nature.com/articles/s41591-026-04521-4}
}

@article{akinwande2026hyperclip,
  title = {Hyper{CLIP}: Prompt-Conditioned Image Encoders for Contrastive Vision-Language Pre-training},
  author = {Akinwande, Victor and Norouzzadeh, Mohammad Sadegh and Willmott, Devin and Bair, Anna and Ganesh, Madan Ravi and Kolter, J Zico},
  journal = {Transactions on Machine Learning Research},
  issn = {2835-8856},
  year = {2026},
  url = {https://openreview.net/forum?id=kX7iwF5s3v}
}

@article{chen2024uni,
  title = {Towards a general-purpose foundation model for computational pathology},
  author = {Chen, Richard J. and Ding, Tong and Lu, Ming Y. and Williamson, Drew F. K. and Jaume, Guillaume and Song, Andrew H. and Chen, Bowen and Zhang, Andrew and Shao, Daniel and Shaban, Muhammad and Williams, Mane and Oldenburg, Lukas and Weishaupt, Luca L. and Wang, Judy J. and Vaidya, Anurag and Le, Long Phi and Gerber, Georg and Sahai, Sharifa and Williams, Walt and Mahmood, Faisal},
  journal = {Nature Medicine},
  volume = {30},
  pages = {850--862},
  year = {2024},
  doi = {10.1038/s41591-024-02857-3},
  url = {https://www.nature.com/articles/s41591-024-02857-3}
}

@article{vorontsov2024virchow,
  title = {A foundation model for clinical-grade computational pathology and rare cancers detection},
  author = {Vorontsov, Eugene and Bozkurt, Alican and Casson, Adam and Shaikovski, George and Zelechowski, Michal and Severson, Kristen and Zimmermann, Eric and Hall, James and Tenenholtz, Neil and Fusi, Nicolo and Yang, Ellen and Mathieu, Philippe and van Eck, Alexander and Lee, Donghun and Viret, Julian and Robert, Eric and Wang, Yi Kan and Kunz, Jeremy D. and Lee, Matthew C. H. and Bernhard, Jan H. and Godrich, Ran A. and Oakley, Gerard and Millar, Ewan and Hanna, Matthew and Wen, Hannah and Retamero, Juan A. and Moye, William A. and Yousfi, Razik and Kanan, Christopher and Klimstra, David S. and Rothrock, Brandon and Liu, Siqi and Fuchs, Thomas J.},
  journal = {Nature Medicine},
  volume = {30},
  pages = {2924--2935},
  year = {2024},
  doi = {10.1038/s41591-024-03141-0},
  url = {https://www.nature.com/articles/s41591-024-03141-0}
}

@article{xu2024gigapath,
  title = {A whole-slide foundation model for digital pathology from real-world data},
  author = {Xu, Hanwen and Usuyama, Naoto and Bagga, Jaspreet and Zhang, Sheng and Rao, Rajesh and Naumann, Tristan and Wong, Cliff and Gero, Zelalem and Gonz{\'a}lez, Javier and Gu, Yu and Xu, Yanbo and Wei, Mu and Wang, Wenhui and Ma, Shuming and Wei, Furu and Yang, Jianwei and Li, Chunyuan and Gao, Jianfeng and Rosemon, Jaylen and Bower, Tucker and Lee, Soohee and Weerasinghe, Roshanthi and Wright, Bill J. and Robicsek, Ari and Piening, Brian and Bifulco, Carlo and Wang, Sheng and Poon, Hoifung},
  journal = {Nature},
  volume = {630},
  pages = {181--188},
  year = {2024},
  doi = {10.1038/s41586-024-07441-w},
  url = {https://www.nature.com/articles/s41586-024-07441-w}
}

@article{wang2024chief,
  title = {A pathology foundation model for cancer diagnosis and prognosis prediction},
  author = {Wang, Xiyue and Zhao, Junhan and Marostica, Eliana and Yuan, Wei and Jin, Jietian and Zhang, Jiayu and Li, Ruijiang and Tang, Hongping and Wang, Kanran and Li, Yu and Wang, Fang and Peng, Yulong and Zhu, Junyou and Zhang, Jing and Jackson, Christopher R. and Zhang, Jun and Dillon, Deborah and Lin, Nancy U. and Sholl, Lynette and Denize, Thomas and Meredith, David and Ligon, Keith L. and Signoretti, Sabina and Ogino, Shuji and Golden, Jeffrey A. and Nasrallah, MacLean P. and Han, Xiao and Yang, Sen and Yu, Kun-Hsing},
  journal = {Nature},
  volume = {634},
  pages = {970--978},
  year = {2024},
  doi = {10.1038/s41586-024-07894-z},
  url = {https://www.nature.com/articles/s41586-024-07894-z}
}

@article{huang2023plip,
  title = {A visual--language foundation model for pathology image analysis using medical {Twitter}},
  author = {Huang, Zhi and Bianchi, Federico and Yuksekgonul, Mert and Montine, Thomas J. and Zou, James},
  journal = {Nature Medicine},
  volume = {29},
  pages = {2307--2316},
  year = {2023},
  doi = {10.1038/s41591-023-02504-3},
  url = {https://www.nature.com/articles/s41591-023-02504-3}
}

@article{campanella2019clinical,
  title = {Clinical-grade computational pathology using weakly supervised deep learning on whole slide images},
  author = {Campanella, Gabriele and Hanna, Matthew G. and Geneslaw, Luke and Miraflor, Allen and Werneck Krauss Silva, Vitor and Busam, Klaus J. and Brogi, Edi and Reuter, Victor E. and Klimstra, David S. and Fuchs, Thomas J.},
  journal = {Nature Medicine},
  volume = {25},
  pages = {1301--1309},
  year = {2019},
  doi = {10.1038/s41591-019-0508-1},
  url = {https://www.nature.com/articles/s41591-019-0508-1}
}

@article{coudray2018classification,
  title = {Classification and mutation prediction from non--small cell lung cancer histopathology images using deep learning},
  author = {Coudray, Nicolas and Ocampo, Paolo Santiago and Sakellaropoulos, Theodore and Narula, Navneet and Snuderl, Matija and Feny{\"o}, David and Moreira, Andre L. and Razavian, Narges and Tsirigos, Aristotelis},
  journal = {Nature Medicine},
  volume = {24},
  pages = {1559--1567},
  year = {2018},
  doi = {10.1038/s41591-018-0177-5},
  url = {https://www.nature.com/articles/s41591-018-0177-5}
}

@article{kather2019msi,
  title = {Deep learning can predict microsatellite instability directly from histology in gastrointestinal cancer},
  author = {Kather, Jakob Nikolas and Pearson, Alexander T. and Halama, Niels and J{\"a}ger, Dirk and Krause, Jeremias and Loosen, Sven H. and Marx, Alexander and Boor, Peter and Tacke, Frank and Neumann, Ulf Peter and Grabsch, Heike I. and Yoshikawa, Takaki and Brenner, Hermann and Chang-Claude, Jenny and Hoffmeister, Michael and Trautwein, Christian and Luedde, Tom},
  journal = {Nature Medicine},
  volume = {25},
  pages = {1054--1056},
  year = {2019},
  doi = {10.1038/s41591-019-0462-y},
  url = {https://www.nature.com/articles/s41591-019-0462-y}
}

@inproceedings{li2021dsmil,
  title = {Dual-Stream Multiple Instance Learning Network for Whole Slide Image Classification With Self-Supervised Contrastive Learning},
  author = {Li, Bin and Li, Yin and Eliceiri, Kevin W.},
  booktitle = {Proceedings of the IEEE/CVF Conference on Computer Vision and Pattern Recognition},
  pages = {14318--14328},
  year = {2021},
  url = {https://openaccess.thecvf.com/content/CVPR2021/html/Li_Dual-Stream_Multiple_Instance_Learning_Network_for_Whole_Slide_Image_Classification_CVPR_2021_paper.html}
}

@inproceedings{zhang2022dtfd,
  title = {{DTFD-MIL}: Double-Tier Feature Distillation Multiple Instance Learning for Histopathology Whole Slide Image Classification},
  author = {Zhang, Hongrun and Meng, Yanda and Zhao, Yitian and Qiao, Yihong and Yang, Xiaoyun and Coupland, Sarah E. and Zheng, Yalin},
  booktitle = {Proceedings of the IEEE/CVF Conference on Computer Vision and Pattern Recognition},
  pages = {18802--18812},
  year = {2022},
  url = {https://openaccess.thecvf.com/content/CVPR2022/html/Zhang_DTFD-MIL_Double-Tier_Feature_Distillation_Multiple_Instance_Learning_for_Histopathology_Whole_CVPR_2022_paper.html}
}

@inproceedings{chen2022hipt,
  title = {Scaling Vision Transformers to Gigapixel Images via Hierarchical Self-Supervised Learning},
  author = {Chen, Richard J. and Chen, Chengkuan and Li, Yicong and Chen, Tiffany Y. and Trister, Andrew D. and Krishnan, Rahul G. and Mahmood, Faisal},
  booktitle = {Proceedings of the IEEE/CVF Conference on Computer Vision and Pattern Recognition},
  pages = {16144--16155},
  year = {2022},
  url = {https://openaccess.thecvf.com/content/CVPR2022/html/Chen_Scaling_Vision_Transformers_to_Gigapixel_Images_via_Hierarchical_Self-Supervised_Learning_CVPR_2022_paper.html}
}

@article{wang2022ctranspath,
  title = {Transformer-based unsupervised contrastive learning for histopathological image classification},
  author = {Wang, Xiyue and Yang, Sen and Zhang, Jun and Wang, Minghui and Zhang, Jing and Yang, Wei and Huang, Junzhou and Han, Xiao},
  journal = {Medical Image Analysis},
  volume = {81},
  pages = {102559},
  year = {2022},
  doi = {10.1016/j.media.2022.102559},
  url = {https://www.sciencedirect.com/science/article/pii/S1361841522002043}
}

@article{wang2023retccl,
  title = {{RetCCL}: Clustering-guided contrastive learning for whole-slide image retrieval},
  author = {Wang, Xiyue and Du, Yuexi and Yang, Sen and Zhang, Jun and Wang, Minghui and Zhang, Jing and Yang, Wei and Huang, Junzhou and Han, Xiao},
  journal = {Medical Image Analysis},
  volume = {83},
  pages = {102645},
  year = {2023},
  doi = {10.1016/j.media.2022.102645},
  url = {https://www.sciencedirect.com/science/article/pii/S1361841522002730}
}

@article{zhang2023biomedclip,
  title = {A Multimodal Biomedical Foundation Model Trained from Fifteen Million Image--Text Pairs},
  author = {Zhang, Sheng and Xu, Yanbo and Usuyama, Naoto and Xu, Hanwen and Bagga, Jaspreet and Tinn, Robert and Preston, Sam and Rao, Rajesh and Wei, Mu and Valluri, Naveen and Wong, Cliff and Tupini, Andrea and Wang, Yu and Mazzola, Matt and Shukla, Swadheen and Liden, Lars and Gao, Jianfeng and Crabtree, Angela and Piening, Brian and Bifulco, Carlo and Lungren, Matthew P. and Naumann, Tristan and Wang, Sheng and Poon, Hoifung},
  journal = {NEJM AI},
  volume = {2},
  number = {1},
  pages = {AIoa2400640},
  year = {2025},
  doi = {10.1056/AIoa2400640},
  url = {https://ai.nejm.org/doi/full/10.1056/AIoa2400640}
}

@article{zimmermann2024virchow2,
  title = {{Virchow2}: Scaling Self-Supervised Mixed Magnification Models in Pathology},
  author = {Zimmermann, Eric and Vorontsov, Eugene and Viret, Julian and Casson, Adam and Zelechowski, Michal and Shaikovski, George and Tenenholtz, Neil and Hall, James and Klimstra, David and Yousfi, Razik and Fuchs, Thomas and Fusi, Nicolo and Liu, Siqi and Severson, Kristen},
  journal = {arXiv preprint arXiv:2408.00738},
  year = {2024},
  url = {https://arxiv.org/abs/2408.00738}
}

@inproceedings{elhoseiny2013write,
  title = {Write a Classifier: Zero-Shot Learning Using Purely Textual Descriptions},
  author = {Elhoseiny, Mohamed and Saleh, Babak and Elgammal, Ahmed},
  booktitle = {Proceedings of the IEEE International Conference on Computer Vision},
  pages = {2584--2591},
  year = {2013},
  url = {https://www.cv-foundation.org/openaccess/content_iccv_2013/html/Elhoseiny_Write_a_Classifier_2013_ICCV_paper.html}
}

@inproceedings{ba2015predicting,
  title = {Predicting Deep Zero-Shot Convolutional Neural Networks Using Textual Descriptions},
  author = {Ba, Jimmy Lei and Swersky, Kevin and Fidler, Sanja and Salakhutdinov, Ruslan},
  booktitle = {Proceedings of the IEEE International Conference on Computer Vision},
  pages = {4247--4255},
  year = {2015},
  url = {https://www.cv-foundation.org/openaccess/content_iccv_2015/html/Ba_Predicting_Deep_Zero-Shot_ICCV_2015_paper.html}
}

@inproceedings{reed2016descriptions,
  title = {Learning Deep Representations of Fine-Grained Visual Descriptions},
  author = {Reed, Scott and Akata, Zeynep and Lee, Honglak and Schiele, Bernt},
  booktitle = {Proceedings of the IEEE Conference on Computer Vision and Pattern Recognition},
  pages = {49--58},
  year = {2016},
  url = {https://openaccess.thecvf.com/content_cvpr_2016/html/Reed_Learning_Deep_Representations_CVPR_2016_paper.html}
}

@article{zhou2022coop,
  title = {Learning to Prompt for Vision-Language Models},
  author = {Zhou, Kaiyang and Yang, Jingkang and Loy, Chen Change and Liu, Ziwei},
  journal = {International Journal of Computer Vision},
  volume = {130},
  number = {9},
  pages = {2337--2348},
  year = {2022},
  doi = {10.1007/s11263-022-01653-1},
  url = {https://link.springer.com/article/10.1007/s11263-022-01653-1}
}

@inproceedings{zhou2022cocoop,
  title = {Conditional Prompt Learning for Vision-Language Models},
  author = {Zhou, Kaiyang and Yang, Jingkang and Loy, Chen Change and Liu, Ziwei},
  booktitle = {Proceedings of the IEEE/CVF Conference on Computer Vision and Pattern Recognition},
  pages = {16816--16825},
  year = {2022},
  url = {https://openaccess.thecvf.com/content/CVPR2022/html/Zhou_Conditional_Prompt_Learning_for_Vision-Language_Models_CVPR_2022_paper.html}
}

@inproceedings{karimimahabadi2021hyperformer,
  title = {Parameter-efficient Multi-task Fine-tuning for Transformers via Shared Hypernetworks},
  author = {Karimi Mahabadi, Rabeeh and Ruder, Sebastian and Dehghani, Mostafa and Henderson, James},
  booktitle = {Proceedings of the 59th Annual Meeting of the Association for Computational Linguistics and the 11th International Joint Conference on Natural Language Processing (Volume 1: Long Papers)},
  pages = {565--576},
  publisher = {Association for Computational Linguistics},
  year = {2021},
  doi = {10.18653/v1/2021.acl-long.47},
  url = {https://aclanthology.org/2021.acl-long.47/}
}

@inproceedings{he2022hyperprompt,
  title = {{HyperPrompt}: Prompt-based Task-Conditioning of Transformers},
  author = {He, Yun and Zheng, Steven and Tay, Yi and Gupta, Jai and Du, Yu and Aribandi, Vamsi and Zhao, Zhe and Li, Yaguang and Chen, Zhao and Metzler, Donald and Cheng, Heng-Tze and Chi, Ed H.},
  booktitle = {Proceedings of the 39th International Conference on Machine Learning},
  series = {Proceedings of Machine Learning Research},
  volume = {162},
  pages = {8678--8690},
  publisher = {PMLR},
  year = {2022},
  url = {https://proceedings.mlr.press/v162/he22f.html}
}

@inproceedings{ivison2023hint,
  title = {{HINT}: Hypernetwork Instruction Tuning for Efficient Zero- and Few-Shot Generalisation},
  author = {Ivison, Hamish and Bhagia, Akshita and Wang, Yizhong and Hajishirzi, Hannaneh and Peters, Matthew},
  booktitle = {Proceedings of the 61st Annual Meeting of the Association for Computational Linguistics (Volume 1: Long Papers)},
  pages = {11272--11288},
  publisher = {Association for Computational Linguistics},
  year = {2023},
  doi = {10.18653/v1/2023.acl-long.631},
  url = {https://aclanthology.org/2023.acl-long.631/}
}

@article{nava2023metaguidance,
  title = {Meta-Learning via Classifier(-free) Diffusion Guidance},
  author = {Nava, Elvis and Kobayashi, Seijin and Yin, Yifei and Katzschmann, Robert K. and Grewe, Benjamin F.},
  journal = {Transactions on Machine Learning Research},
  issn = {2835-8856},
  year = {2023},
  url = {https://openreview.net/forum?id=1irVjE7A3w}
}

@inproceedings{zaheer2017deepsets,
  title = {Deep Sets},
  author = {Zaheer, Manzil and Kottur, Satwik and Ravanbakhsh, Siamak and Poczos, Barnabas and Salakhutdinov, Ruslan and Smola, Alexander J.},
  booktitle = {Advances in Neural Information Processing Systems},
  volume = {30},
  pages = {3391--3401},
  year = {2017},
  url = {https://proceedings.neurips.cc/paper/2017/hash/f22e4747da1aa27e363d86d40ff442fe-Abstract.html}
}

@inproceedings{lee2019settransformer,
  title = {Set Transformer: A Framework for Attention-based Permutation-Invariant Neural Networks},
  author = {Lee, Juho and Lee, Yoonho and Kim, Jungtaek and Kosiorek, Adam and Choi, Seungjin and Teh, Yee Whye},
  booktitle = {Proceedings of the 36th International Conference on Machine Learning},
  series = {Proceedings of Machine Learning Research},
  volume = {97},
  pages = {3744--3753},
  publisher = {PMLR},
  year = {2019},
  url = {https://proceedings.mlr.press/v97/lee19d.html}
}

@inproceedings{loshchilov2019adamw,
  title = {Decoupled Weight Decay Regularization},
  author = {Loshchilov, Ilya and Hutter, Frank},
  booktitle = {International Conference on Learning Representations},
  year = {2019},
  url = {https://openreview.net/forum?id=Bkg6RiCqY7}
}

@inproceedings{ling2024amdmil,
  title = {Agent Aggregator with Mask Denoise Mechanism for Histopathology Whole Slide Image Analysis},
  author = {Ling, Xitong and Ouyang, Minxi and Wang, Yizhi and Chen, Xinrui and Yan, Renao and Chu, Hongbo and Cheng, Junru and Guan, Tian and Tian, Sufang and Liu, Xiaoping and He, Yonghong},
  booktitle = {Proceedings of the 32nd ACM International Conference on Multimedia},
  year = {2024},
  pages = {2795--2803},
  publisher = {ACM},
  doi = {10.1145/3664647.3681425},
  url = {https://doi.org/10.1145/3664647.3681425}
}

@inproceedings{fu2026deformablegraph,
  title = {Deformable Attention Graph Representation Learning for Histopathology Whole Slide Image Analysis},
  author = {Fu, Mingxi and Ling, Xitong and Chen, Yuxuan and Li, Jiawen and Fu, Fanglei and Yuan, Huaitian and Guan, Tian and He, Yonghong and Zhu, Lianghui},
  booktitle = {ICASSP 2026 - 2026 IEEE International Conference on Acoustics, Speech and Signal Processing (ICASSP)},
  year = {2026},
  pages = {6471--6475},
  publisher = {IEEE},
  doi = {10.1109/ICASSP55912.2026.11464653},
  url = {https://doi.org/10.1109/ICASSP55912.2026.11464653}
}

@article{yan2025pathorchestra,
  title = {{PathOrchestra}: a comprehensive foundation model for computational pathology with over 100 diverse clinical-grade tasks},
  author = {Yan, Fang and Wu, Jianfeng and Li, Jiawen and Wang, Wei and Chen, Yirong and Wei, Linda and Lu, Jiaxuan and Chen, Wen and Gao, Zizhao and Li, Jianan and Li, Heng and Yan, Hong and Ma, Jiabo and Chen, Minda and Lu, Yang and Chen, Qing and Wang, Yizhi and Ling, Xitong and Wang, Xuenian and Wang, Zihan and Huang, Qiang and Hua, Shengyi and Liu, Mianxin and Ma, Lei and Shen, Tian and Zhang, Xiaofan and He, Yonghong and Chen, Hao and Zhang, Shaoting and Wang, Zhe},
  journal = {npj Digital Medicine},
  volume = {8},
  number = {1},
  pages = {695},
  year = {2025},
  doi = {10.1038/s41746-025-02027-w},
  url = {https://www.nature.com/articles/s41746-025-02027-w}
}

@article{zhu2026digepath,
  title = {Subspecialty-specific foundation model for intelligent gastrointestinal pathology},
  author = {Zhu, Lianghui and Ling, Xitong and Ouyang, Minxi and Liu, Xiaoping and Guan, Tian and Fu, Mingxi and Zeng, Maomao and Cheng, Zhiqiang and Fu, Fanglei and Huang, Qiang and Zhu, Mingxi and Jin, Yibo and He, Qiming and Wang, Yizhi and Cheng, Junru and Wang, Xuanyu and Xie, Luxi and Li, Houqiang and Tian, Sufang and He, Yonghong},
  journal = {npj Digital Medicine},
  volume = {9},
  number = {1},
  pages = {473},
  year = {2026},
  doi = {10.1038/s41746-026-02684-5},
  url = {https://www.nature.com/articles/s41746-026-02684-5}
}

@article{ling2025camelyonplus,
  author = {Ling, Xitong and Lei, Yuanyuan and Li, Jiawen and Cheng, Junru and Huang, Wenting and Guan, Tian and Guan, Jian and He, Yonghong},
  title = {{Comprehensive Benchmark Dataset for Pathological Lymph Node Metastasis in Breast Cancer Sections}},
  journal = {Scientific Data},
  volume = {12},
  number = {1},
  pages = {1381},
  year = {2025},
  doi = {10.1038/s41597-025-05586-5},
  url = {https://www.nature.com/articles/s41597-025-05586-5}
}

@inproceedings{chen2026beyondvit,
  author = {Chen, Weiming and Ling, Xitong and Cai, Zhenyang and Wang, Xidong and Li, Jiawen and Guan, Tian and Wang, Benyou and He, Yonghong},
  title = {{Beyond ViT Tokens}: Masked-Diffusion Pretrained Convolutional Pathology Foundation Model for Cell-Level Dense Prediction},
  booktitle = {Conference on Neural Information Processing Systems},
  year = {2026},
  note = {Poster},
  url = {https://nips.cc/virtual/2026/poster/148943}
}

@inproceedings{liu2026steph,
  title = {Sparse Task Vector Mixup with Hypernetworks for Efficient Knowledge Transfer in Whole-Slide Image Prognosis},
  author = {Liu, Pei and Zeng, Xiangxiang and Ma, Tengfei and Xing, Yucheng and Ren, Xuanbai and Liu, Yiping},
  booktitle = {Proceedings of the IEEE/CVF Conference on Computer Vision and Pattern Recognition},
  year = {2026},
  pages = {35238--35247},
  url = {https://openaccess.thecvf.com/content/CVPR2026/html/Liu_Sparse_Task_Vector_Mixup_with_Hypernetworks_for_Efficient_Knowledge_Transfer_CVPR_2026_paper.html}
}

@inproceedings{li2024wikg,
  author = {Li, Jiawen and Chen, Yuxuan and Chu, Hongbo and Sun, Qiehe and Guan, Tian and Han, Anjia and He, Yonghong},
  title = {Dynamic Graph Representation with Knowledge-aware Attention for Histopathology Whole Slide Image Analysis},
  booktitle = {Proceedings of the IEEE/CVF Conference on Computer Vision and Pattern Recognition},
  year = {2024},
  pages = {11323--11332},
  doi = {10.1109/CVPR52733.2024.01076},
  url = {https://openaccess.thecvf.com/content/CVPR2024/html/Li_Dynamic_Graph_Representation_with_Knowledge-aware_Attention_for_Histopathology_Whole_Slide_CVPR_2024_paper.html}
}

@inproceedings{javed2024cplip,
  author = {Javed, Sajid and Mahmood, Arif and Ganapathi, Iyyakutti Iyappan and Dharejo, Fayaz Ali and Werghi, Naoufel and Bennamoun, Mohammed},
  title = {{CPLIP}: Zero-Shot Learning for Histopathology with Comprehensive Vision-Language Alignment},
  booktitle = {Proceedings of the IEEE/CVF Conference on Computer Vision and Pattern Recognition},
  year = {2024},
  pages = {11450--11459},
  url = {https://openaccess.thecvf.com/content/CVPR2024/html/Javed_CPLIP_Zero-Shot_Learning_for_Histopathology_with_Comprehensive_Vision-Language_Alignment_CVPR_2024_paper.html}
}

@inproceedings{babu2024hyperfields,
  title = {{HyperFields}: Towards Zero-Shot Generation of {NeRFs} from Text},
  author = {Babu, Sudarshan and Liu, Richard and Zhou, Avery and Maire, Michael and Shakhnarovich, Greg and Hanocka, Rana},
  booktitle = {Proceedings of the 41st International Conference on Machine Learning},
  series = {Proceedings of Machine Learning Research},
  volume = {235},
  pages = {2230--2247},
  publisher = {PMLR},
  year = {2024},
  url = {https://proceedings.mlr.press/v235/babu24a.html}
}

}

\clearpage

\appendix
\onecolumn
\setlength{\columnsep}{26pt}

\setcounter{section}{0}
\setcounter{figure}{0}
\setcounter{table}{0}
\renewcommand{\thesection}{S\arabic{section}}
\renewcommand{\thefigure}{S\arabic{figure}}
\renewcommand{\thetable}{S\arabic{table}}
\renewcommand{\theHsection}{supp.\arabic{section}}
\renewcommand{\theHfigure}{supp.\arabic{figure}}
\renewcommand{\theHtable}{supp.\arabic{table}}
\setcounter{page}{1}
\renewcommand{\thepage}{S\arabic{page}}
\section*{Supplementary Material}
\begin{multicols}{2}
\section{Datasets and Task Distribution}
\label{sec:supplement}
\label{sec:registry-distributions}
A task is a dataset--endpoint pair, so the same endpoint in different datasets defines distinct tasks. Figure~\ref{fig:registry-overview} groups the tasks by anatomy, endpoint type and dataset source. Mutation prediction and staging account for 122 and 69 tasks, respectively. Colorectal tasks form the largest anatomical group, with 37 tasks; lymph node and thymus each contain one. These unequal group sizes motivate the endpoint, anatomy and source summaries in the main text.

All comparisons use the same 308 tasks and complete query records. MI-Hyper uses the fixed five-fold checkpoints for 112 semantic groups. The registry contains 160,989 task--bag records and 29,481 recorded patient IDs; task--bag counts can repeat cases that support multiple endpoints. The 52 dataset entries include merged collections and subsets and therefore do not represent 52 independent clinical cohorts.
\end{multicols}

\par
\noindent\begin{minipage}{\textwidth}
\captionsetup{type=figure}
\centering
\includegraphics[width=.96\textwidth]{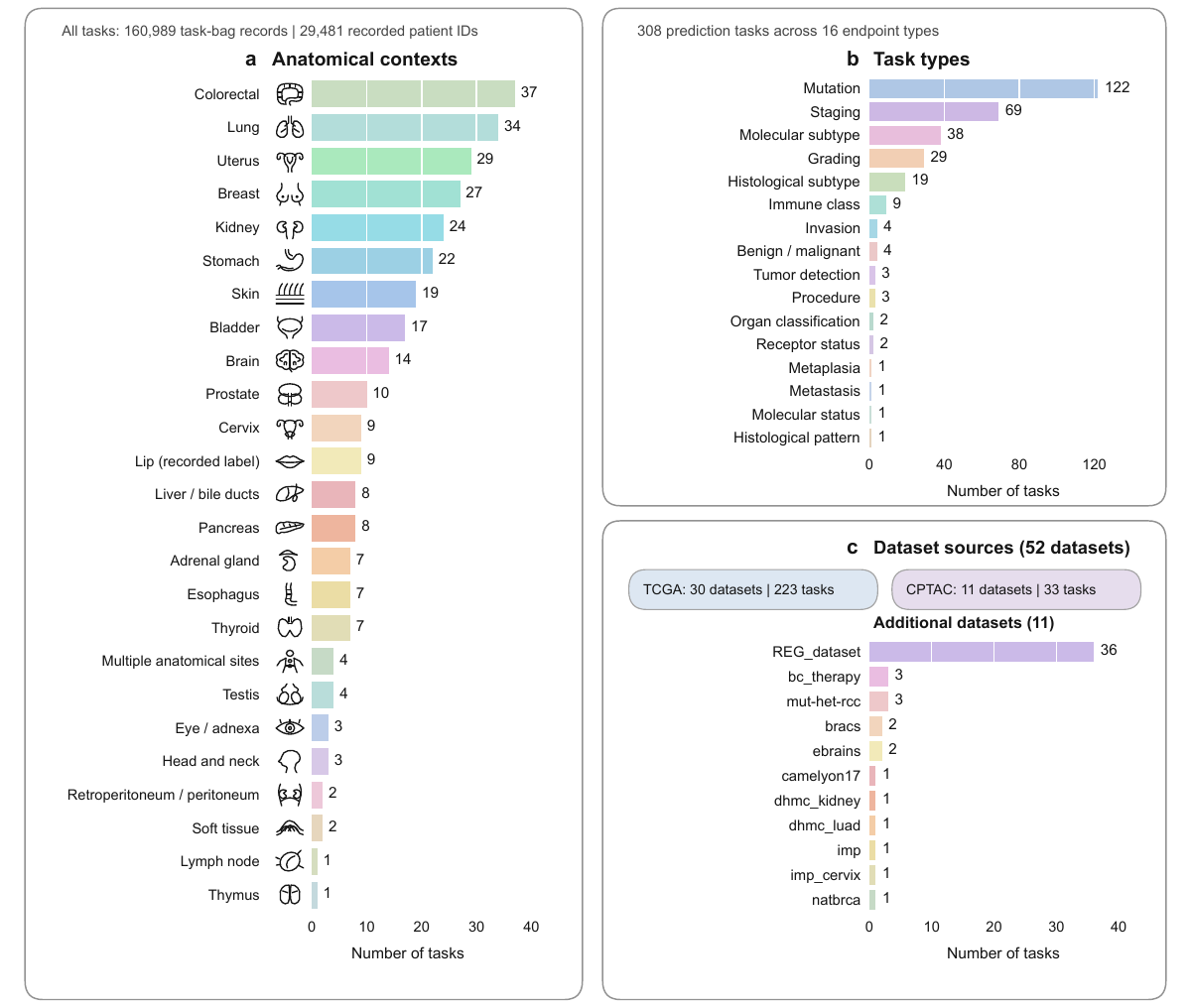}
\caption{\textbf{Composition of the evaluation tasks.} (a) Task counts in 25 anatomical display groups formed by merging the registry's 29 anatomical labels. (b) All 308 tasks grouped into 16 endpoint types. (c) All 52 pathology datasets, including 11 additional datasets listed individually. Counts are taken from the task registry and include every group. Anatomical icons are schematic.}
\label{fig:registry-overview}
\end{minipage}\par

\clearpage
\begin{multicols}{2}
Figure~\ref{fig:dataset-task-catalogue} shows which endpoints are represented in each dataset: TCGA contributes 223 tasks, CPTAC 33 and the additional datasets 52. Because endpoint composition differs across sources, the main-text source averages reflect different task collections. The catalogue uses the same dataset and task identifiers as Table~\ref{tab:si-v18-all308}, linking each group mean to its constituent endpoints.

For anatomical displays and performance summaries, we merge colon, colorectal and colorectum as Colorectal; prostate and prostate gland as Prostate; and cervix and uterine cervix as Cervix. These display-only merges leave original task IDs, anatomical labels and training and evaluation assignments unchanged.
\end{multicols}

\par
\noindent\begin{minipage}{\textwidth}
\captionsetup{type=figure}
\centering
\includegraphics[width=\textwidth]{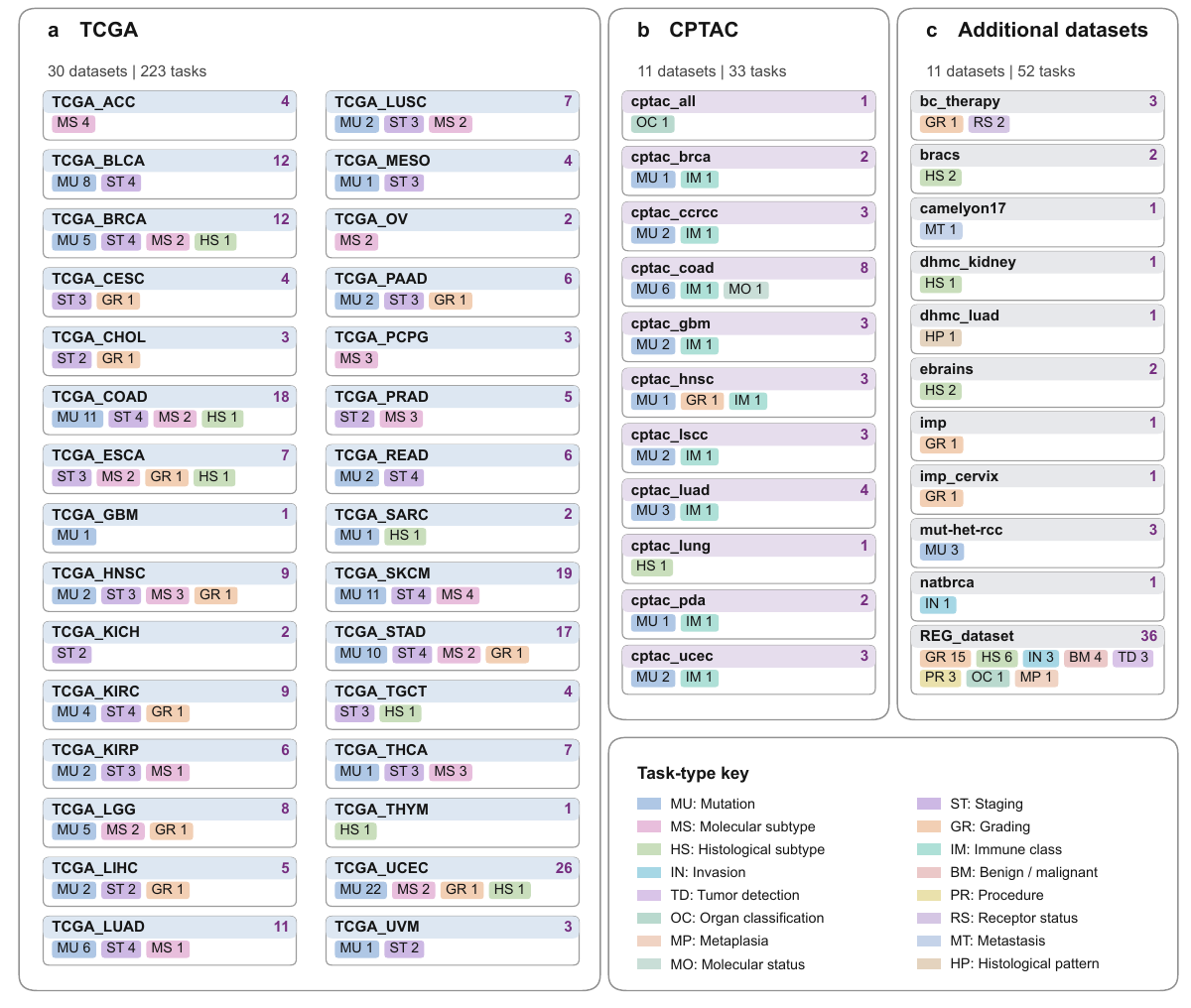}
\caption{\textbf{Endpoint coverage by dataset.} (a) 30 TCGA datasets and 223 tasks. (b) 11 CPTAC datasets and 33 tasks. (c) 11 additional datasets and 52 tasks. Each card identifies a dataset and its task count; colored tags give nonzero counts for each endpoint type. All 52 datasets and 308 tasks are retained, including registered merged collections and subsets. Card area does not encode sample size.}
\label{fig:dataset-task-catalogue}
\end{minipage}\par

\clearpage

\begin{multicols}{2}
\subsection{Attention Maps Across Tasks}
\label{sec:si-task-attention}
Figure~\ref{fig:si-task-attention} pairs H\&E images with task-specific attention maps for 16 tasks from six TCGA datasets. Each pair uses the same WSI and field of view, with the fixed models, complete patch sets and seven-condition averaging of the main evaluation. Slides are chosen by fixed patient-ordering and file-availability rules.

These panels show different tasks and WSIs; main-text Figure~\ref{fig:tcga-attention}c compares tasks on the same slide. Region-level pathology annotations were unavailable for these examples.
\end{multicols}

\par
\noindent\begin{minipage}{\textwidth}
\captionsetup{type=figure}
\centering
\includegraphics[page=1,width=.98\textwidth]{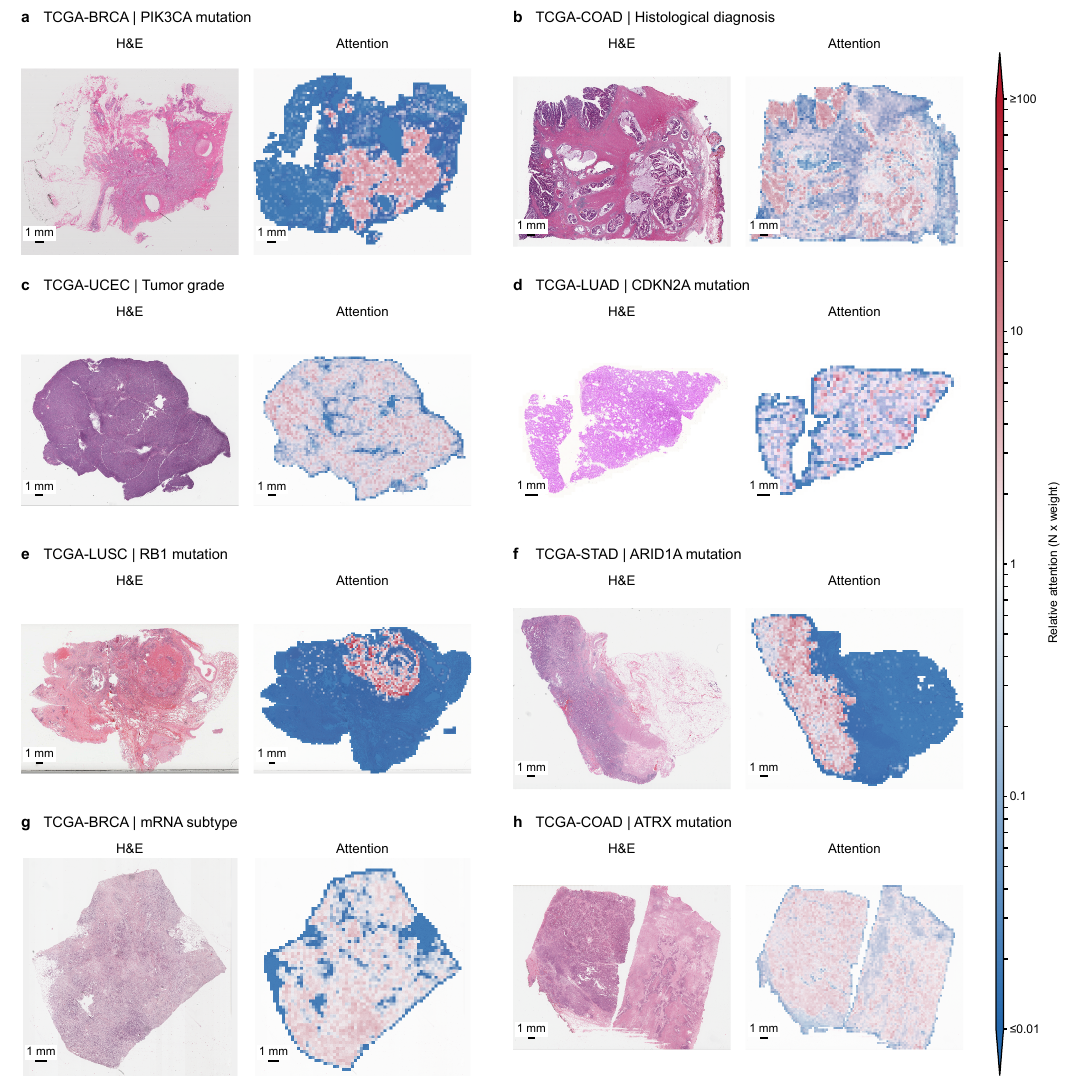}
\caption{\textbf{H\&E images and attention maps across tasks.} Each panel pairs H\&E (left) and attention (right) for the same WSI field. Panels (a)--(h) appear here; (i)--(p) continue on the next page. Maps average seven separately normalized attention distributions. Colors show relative attention, $N\bar{\alpha}$, on a logarithmic scale (0.01--100): blue is lower, red higher. Scale bars: 1\,mm.}
\label{fig:si-task-attention}
\end{minipage}\par

\clearpage
\par
\noindent\begin{minipage}{\textwidth}
\captionsetup{type=figure}
\ContinuedFloat
\centering
\includegraphics[page=2,width=.98\textwidth]{MI-Hyper_Different_Tasks_Attention.pdf}
\caption{\textbf{H\&E images and attention maps across tasks (continued).} Panels (i)--(p). H\&E images and corresponding attention maps use matched fields of view. Text-condition averaging, color scale and scale bars follow panels (a)--(h).}
\end{minipage}\par

\clearpage
\begin{multicols}{2}
\section{Complete-Bag Inference}
\label{sec:si-inference-details}
For the gated implementation, the cached task-specific parameters contain
\begin{equation}
 a+C_\tau(h+1)=128+257C_\tau
 \label{eq:mihyper-cache}
\end{equation}
scalars. Their number depends on the class vocabulary rather than bag length. The shared basis, pretrained encoders and patch features are stored separately.

To apply the predictor to every stored patch, we process long bags in chunks while maintaining a single global attention normalization. Let $m$, $s$ and $u$ denote the running maximum score, normalization sum and weighted feature sum. For a new chunk $\mathcal B$, the updates are
\begin{align}
 m'&=\max\left\{m,\max_{i\in\mathcal B}r_i\right\},\nonumber\\
 s'&=\exp(m-m')s+
 \sum_{i\in\mathcal B}\exp(r_i-m'),\nonumber\\
 u'&=\exp(m-m')u+
 \sum_{i\in\mathcal B}\exp(r_i-m')v_i.
 \label{eq:streaming}
\end{align}
The state is initialized from the first chunk, and the final aggregated representation is $v_X=u/s$. The recurrence uses all stored patches for globally normalized pooling, up to floating-point rounding. Training samples at most 2048 patches per bag to control cost; chunking instead executes each complete query bag.
\end{multicols}

\definecolor{AllTasksTint}{HTML}{F9E9EF}
\definecolor{HyperTint}{HTML}{EAF3FA}
\definecolor{SIGray}{HTML}{6D6D6D}
\definecolor{SIPurple}{HTML}{762E80}
\newlength{\SIContentWidth}
\newlength{\SINumberWidth}
\newcommand{\SIStyle}{%
\fontsize{8}{9.2}\selectfont
\setlength{\tabcolsep}{1.6pt}%
\renewcommand{\arraystretch}{1.08}%
\setlength{\heavyrulewidth}{0.8pt}%
\setlength{\lightrulewidth}{0.4pt}%
\setlength{\aboverulesep}{1.6pt}%
\setlength{\belowrulesep}{1.6pt}%
\setlength{\LTcapwidth}{\textwidth}%
\setlength{\LTpre}{\smallskipamount}%
\setlength{\LTpost}{\smallskipamount}%
}
\newcommand{\SIBase}[1]{\textcolor{SIGray}{#1}}
\newcommand{\SIHeader}[1]{\textcolor{SIPurple}{#1}}

\begin{multicols}{2}
\section{Results for All 308 Tasks}
\label{sec:all-current-results}
\label{sec:si-v18-text}
Table~\ref{tab:si-v18-all308} extends the grouped AUC and BAcc comparisons in the main text to AUC, BAcc and macro-F1 for every task. MI-Zero includes five pooling settings and Mean-$K$; MI-Hyper averages metrics across seven predefined conditions. Both methods use the same query list for each task, and MI-Hyper uses its fixed model without target-label selection of text conditions. Class and bag counts are calculated from this query list.

MI-Hyper averages 58.24\% AUC, 45.66\% BAcc and 36.05\% F1 across tasks, compared with 56.38\%, 44.49\% and 35.48\% for MI-Zero. The table includes results favoring either method, so readers can relate the overall and grouped means to individual endpoints. Blue task IDs mark the six main-text examples with concurrent gains; this display subset does not change the overall statistics.
\end{multicols}

\par
\begingroup\SIStyle
\setlength{\SIContentWidth}{\dimexpr\textwidth-20\tabcolsep\relax}
\setlength{\SINumberWidth}{\dimexpr(\SIContentWidth-1.82in-.30in-.13in-.33in)/7\relax}
\begin{longtable}{@{}>{\raggedright\arraybackslash}m{1.82in}>{\raggedleft\arraybackslash}m{.30in}>{\raggedleft\arraybackslash}m{.13in}>{\centering\arraybackslash}m{.33in}>{\centering\arraybackslash}m{\SINumberWidth}>{\centering\arraybackslash}m{\SINumberWidth}>{\centering\arraybackslash}m{\SINumberWidth}>{\centering\arraybackslash}m{\SINumberWidth}>{\centering\arraybackslash}m{\SINumberWidth}>{\centering\arraybackslash}m{\SINumberWidth}>{\columncolor{HyperTint}[\tabcolsep][0pt]\centering\arraybackslash}m{\SINumberWidth}@{}}
\caption{\textbf{All 308 tasks under the same 112-group partition.} Each cell lists AUC, BAcc and macro-F1 (\%) from top to bottom. Each MI-Zero $K$ column averages metrics over 50 prompts; Mean-$K$ averages all 250 conditions. MI-Hyper uses the fixed five-fold checkpoints and averages metrics over seven predefined paired text conditions. All columns use metric averaging, without selecting the best condition for each task. $N$ is the number of task--bag records and $C$ the number of classes. Blue task IDs mark the six main-text examples; MI-Zero scores are gray.}\label{tab:si-v18-all308}\label{tab:all308-measured}\\
\toprule

\multicolumn{4}{@{}l}{} & \multicolumn{6}{c}{MI-Zero} & \multicolumn{1}{>{\columncolor{HyperTint}[\tabcolsep][0pt]}c@{}}{\SIHeader{MI-Hyper}}\\

Task & $N$ & $C$ & Metric & \shortstack{$K=$\\1} & \shortstack{$K=$\\5} & \shortstack{$K=$\\10} & \shortstack{$K=$\\50} & \shortstack{$K=$\\100} & \shortstack{Mean\\$K$} & \SIHeader{\shortstack{7-condition\\mean}} \\
\midrule
\endfirsthead
\multicolumn{11}{@{}l}{\textcolor{SIGray}{\tablename\ \thetable\ (continued)}}\\
\toprule

\multicolumn{4}{@{}l}{} & \multicolumn{6}{c}{MI-Zero} & \multicolumn{1}{>{\columncolor{HyperTint}[\tabcolsep][0pt]}c@{}}{\SIHeader{MI-Hyper}}\\

Task & $N$ & $C$ & Metric & \shortstack{$K=$\\1} & \shortstack{$K=$\\5} & \shortstack{$K=$\\10} & \shortstack{$K=$\\50} & \shortstack{$K=$\\100} & \shortstack{Mean\\$K$} & \SIHeader{\shortstack{7-condition\\mean}} \\
\midrule
\endhead
\midrule
\multicolumn{11}{@{}r}{\textcolor{SIGray}{Continued on next page}}\\
\endfoot
\bottomrule
\endlastfoot
\multicolumn{10}{@{}l}{\textcolor{SIPurple}{\bfseries REG\_dataset}} & \cellcolor{HyperTint}{}\\*
\texttt{REG\_\allowbreak{}dataset/\allowbreak{}bladder\_\allowbreak{}grade} & 348 & 2 & \textcolor{SIGray}{\shortstack{AUC\\BAcc\\F1}} & \SIBase{\shortstack{88.95\\64.26\\54.12}} & \SIBase{\shortstack{90.70\\63.84\\53.41}} & \SIBase{\shortstack{91.02\\63.73\\53.26}} & \SIBase{\shortstack{90.31\\62.14\\50.72}} & \SIBase{\shortstack{89.55\\60.37\\47.81}} & \SIBase{\shortstack{90.11\\62.87\\51.87}} & \shortstack{90.39\\50.69\\39.43} \\
\cmidrule[0.15pt](lr){1-11}
\texttt{REG\_\allowbreak{}dataset/\allowbreak{}bladder\_\allowbreak{}invasion\_\allowbreak{}depth} & 867 & 3 & \textcolor{SIGray}{\shortstack{AUC\\BAcc\\F1}} & \SIBase{\shortstack{75.40\\43.41\\39.63}} & \SIBase{\shortstack{79.35\\44.83\\40.95}} & \SIBase{\shortstack{79.83\\44.18\\40.32}} & \SIBase{\shortstack{77.22\\41.08\\36.66}} & \SIBase{\shortstack{75.26\\38.40\\33.22}} & \SIBase{\shortstack{77.41\\42.38\\38.15}} & \shortstack{74.77\\51.68\\47.33} \\
\cmidrule[0.15pt](lr){1-11}
\texttt{REG\_\allowbreak{}dataset/\allowbreak{}bladder\_\allowbreak{}invasiveness} & 808 & 3 & \textcolor{SIGray}{\shortstack{AUC\\BAcc\\F1}} & \SIBase{\shortstack{87.76\\64.98\\59.20}} & \SIBase{\shortstack{91.47\\66.20\\61.56}} & \SIBase{\shortstack{92.62\\67.15\\62.21}} & \SIBase{\shortstack{94.12\\68.84\\63.83}} & \SIBase{\shortstack{93.88\\71.05\\64.38}} & \SIBase{\shortstack{91.97\\67.65\\62.24}} & \shortstack{83.45\\42.65\\37.06} \\
\cmidrule[0.15pt](lr){1-11}
\texttt{REG\_\allowbreak{}dataset/\allowbreak{}bladder\_\allowbreak{}tumor\_\allowbreak{}presence} & 867 & 2 & \textcolor{SIGray}{\shortstack{AUC\\BAcc\\F1}} & \SIBase{\shortstack{53.66\\51.45\\30.65}} & \SIBase{\shortstack{60.84\\55.36\\30.20}} & \SIBase{\shortstack{63.98\\57.07\\30.41}} & \SIBase{\shortstack{69.56\\61.13\\31.32}} & \SIBase{\shortstack{70.69\\61.36\\31.23}} & \SIBase{\shortstack{63.75\\57.27\\30.76}} & \shortstack{28.24\\31.10\\20.34} \\
\end{longtable}\endgroup

\end{document}